%% file: main.tex
\documentclass[letterpaper, 10 pt, conference]{ieeeconf}  

\IEEEoverridecommandlockouts                              

\usepackage{algpseudocode} 
\usepackage{amsmath,amsfonts}
\usepackage{amssymb}
\usepackage{mathrsfs}
\usepackage{algorithm}
\usepackage{array}
\usepackage{textcomp}
\usepackage{soul}
\usepackage{color,xcolor}
\usepackage{stfloats}
\usepackage{url}
\usepackage{verbatim}
\usepackage{graphicx}
\usepackage{cite}

\usepackage{amsthm}
\usepackage{cleveref}
\usepackage{makecell}
\usepackage{multirow}
\usepackage{placeins}
\usepackage{mathtools}
\usepackage{bm}

\usepackage{subcaption}
\usepackage{lipsum}
\usepackage{stfloats}
\usepackage{booktabs}

\usepackage{wrapfig}

\usepackage{pifont}
\newcommand{\cmark}{\textcolor{green!60!black}{\ding{51}}}
\newcommand{\xmark}{\textcolor{red}{\ding{55}}}

\definecolor{editblue}{rgb}{0.1, 0.4, 0.85}
\definecolor{comment}{rgb}{0, 0, 0}

\usepackage{xcolor}    
\usepackage{xpatch}    
\usepackage{pgffor}
\makeatletter
\newcommand{\bibColoredItems}[2]{%
    \foreach \x in {#2} {%
        \expandafter\gdef\csname bib@colored@\x\endcsname{#1}%
    }%
}

\newcommand\bib@setcolor[1]{%
  \ifcsname bib@colored@#1\endcsname
    \expandafter\color\expandafter{\csname bib@colored@#1\endcsname}
  \else
    \normalcolor
  \fi
}

\IfPackageLoadedTF{hyperref}{\@tempswatrue}{\@tempswafalse}
\if@tempswa
  \xpatchcmd\@bibitem{\H@item}{\bib@setcolor{#1}\H@item}{}{\PatchFailed}
  \xpatchcmd\@lbibitem{\H@item}{\bib@setcolor{#2}\H@item}{}{\PatchFailed}
\else
  \xpatchcmd\@bibitem{\item}{\bib@setcolor{#1}\item}{}{\PatchFailed}
  \xpatchcmd\@lbibitem{\item}{\bib@setcolor{#2}\item}{}{\PatchFailed}
\fi
\makeatother

\title{\LARGE \bf
Dual-Quadruped Collaborative Transportation in Narrow Environments via Safe Reinforcement Learning
}

\author{Zhezhi Lei, Zhihai Bi, Wenxin Wang, and Jun Ma, \textit{Senior Member, IEEE}
 \thanks{Zhezhi Lei and Zhihai Bi are with the Robotics and Autonomous Systems Thrust, The Hong Kong University of Science and Technology (Guangzhou), Guangzhou 511453, China (e-mail: zhezhilei@hkust-gz.edu.cn; zbi217@connect.hkust-gz.edu.cn). }
 \thanks{Wenxin Wang is with Advanced Remanufacturing \& Technology Centre (ARTC), Agency for Science, Technology and Research (A*STAR), Singapore 637143 (e-mail: wang\_wenxin@a-star.edu.sg).} 
\thanks{Jun Ma is with the Robotics and Autonomous Systems Thrust, The Hong Kong University of Science and Technology (Guangzhou), Guangzhou 511453, China, and also with the Cheng Kar-Shun Robotics Institute, The Hong Kong University of Science and Technology, Hong Kong SAR, China (e-mail: jun.ma@ust.hk).} 
}

\begin{document}

\maketitle
\thispagestyle{empty}
\pagestyle{empty}

\input{0Abstract/abstract}
\input{1Introduction/introduction}

\input{2ProblemStatement/problem_statement}
\input{3ProblemSolving/problem_solving}

\input{4Simulation/simulation}
\input{5Conclusion/conclusion}

\bibColoredItems{editblue}{chen2021constrained}
\bibColoredItems{editblue}{chen2021constrained2}
\bibColoredItems{editblue}{ren2024safe}
\bibColoredItems{editblue}{dai2019chance}
\bibColoredItems{editblue}{lyons2012chance}
\bibColoredItems{editblue}{chen2017constrained}
\bibColoredItems{editblue}{shimizu2020motion}

\bibliographystyle{IEEEtran}
\bibliography{ref}

\end{document}

%% file: 0Abstract/abstract.tex
\begin{abstract}
Collaborative transportation, where multiple robots collaboratively transport a payload, has garnered significant attention in recent years. While ensuring safe and high-performance inter-robot collaboration is critical for effective task execution, it is difficult to pursue in narrow environments where the feasible region is extremely limited. To address this challenge, we propose a novel approach for dual-quadruped collaborative transportation via safe reinforcement learning (RL). Specifically, we model the task as a fully cooperative constrained Markov game, where collision avoidance is formulated as constraints. We introduce a cost–advantage decomposition method that enforces the sum of team constraints to remain below an upper bound, thereby guaranteeing task safety within an RL framework. Furthermore, we propose a constraint allocation method that assigns shared constraints to individual robots to maximize the overall task reward, encouraging autonomous task-assignment among robots, thereby improving collaborative task performance. Simulation and real-time experimental results demonstrate that the proposed approach achieves superior performance and a higher success rate in dual-quadruped collaborative transportation compared to existing methods.
\end{abstract}


%% file: 1Introduction/introduction.tex
\section{INTRODUCTION}

Collaborative transportation with robots 
has demonstrated significant potential in applications such as search-and-rescue \cite{tian2025multi}, construction \cite{krizmancic2020cooperative}, and logistics \cite{nie2024social}. 
In these applications, decentralized architectures have attracted increasing attention for their low communication cost, as each robot does not need to exchange data with a centralized controller. 
However, such systems pose substantial challenges in safety assurance and inter-robot coordination, 
as each robot must simultaneously make decisions to avoid potential collisions in narrow environments and collaborate with other robots for successful task execution.

Typical solutions to decentralized collaborative transportation include the model-based method \cite{kim2023layered,lei2025safe, imran2025safety} and the learning-based method \cite{shibata2021deep,ji2021reinforcement,nie2024predictive}. 
For model-based methods, dynamic and collision constraints are incorporated to ensure solution feasibility, 
and optimal solutions are obtained using techniques such as distributed Model Predictive Control \cite{sundin2022decentralized}. 
These methods provide strong safety and feasibility guarantees by explicitly incorporating system dynamics and collision constraints into optimization problems. 
However, model-based approaches are sensitive to the accuracy of the model information. 
In contrast, learning-based methods learn policies directly from interaction data without relying on explicit system models. 
One of the most representative approaches is reinforcement learning (RL). 
Jose et al. \cite{jose2024bilevel} propose a bilevel RL method for dual-quadruped transportation, achieving superior motion planning performance over conventional approaches.
Feng et al. \cite{feng2025learning} develop a hierarchical multi-agent RL framework for complex task planning.
While RL-based methods have shown promising performance in collaborative transportation, decentralized learning frameworks face severe non-stationarity issues, 
since independent policy updates break the stationarity assumption necessary for convergence and compromise learning stability.

To address these challenges,   
actor-critic based RL methods \cite{han2020actor} utilize a value function estimator to reduce variance and improve sample efficiency.  
But large update steps may still cause sudden performance drops, posing risks in safety-critical settings. 
To address this concern, trust-region learning \cite{schulman2015trust} constrains updates within a local region, ensuring monotonic performance improvement. 
Building on this idea, MAPPO \cite{yu2022surprising} applies clipping independently to each agent to prevent overly large steps. 
Nevertheless, MAPPO updates all robots synchronously, which still induces non-stationarity and destabilizes training. 
To further improve stability, HAPPO \cite{kuba2022trust} proposes a sequential update scheme which implicitly encourages collaboration, ultimately enabling superior stability and better performance in collaborative tasks.
However, HAPPO ignores task constraints, making it challenging to satisfy safety requirements in narrow environments, where the feasible region for multi-robot systems is highly constrained.

Considering safety performance in narrow environments, Cao et al. \cite{cao2024hma} incorporate penalty terms to mitigate unsafe behaviors.
However, such approaches do not offer strict safety guarantees. 
To address this limitation, projection-based methods \cite{lin2024projection,razmjoo2024sampling} have been introduced, which project actions onto the feasible set to enforce safety.  
However, they do not directly learn safe policies. 
As a result, frequent projections may cause significant deviations between executed actions and the intended policy, severely degrading performance. 
To overcome this issue, MAPPO-Lagrangian \cite{gu2023safe} incorporates independent safety constraints for each robot based on the HAPPO framework. 
Nevertheless, the independent-constraint assumption makes such methods struggle to handle task constraints requiring high-performance inter-robot collaboration, 
since physical attachments to the payload cause each robot’s actions to significantly influence the states and action spaces of others.

To cater to the aforementioned requirements for decentralized collaborative tasks, 
a classic method is to manually assign leader-follower roles to structure collaboration \cite{sui2020formation}. 
Nevertheless, fixed roles lack flexibility and fail to adapt to varying environments. 
To overcome this limitation, role-learning methods \cite{hou2023multiagent} improve autonomy and adaptability by explicit role-assignment strategies, where each robot maintains a role policy and acts based on its chosen role. 
However, it still requires predefined role types and prior knowledge of coordination needs. 
Alternatively, implicit role-assignment approaches \cite{grannen2023stabilize} maximize policy diversity, 
encouraging robots to spontaneously develop complementary behaviors without explicit role labels. 
However, due to its lack of task-oriented constraints, the assigned roles may misalign with task objectives, 
ultimately limiting stable collaborative control. 

\begin{figure*}[t]
    \centering
    \includegraphics[width=1.0\textwidth]{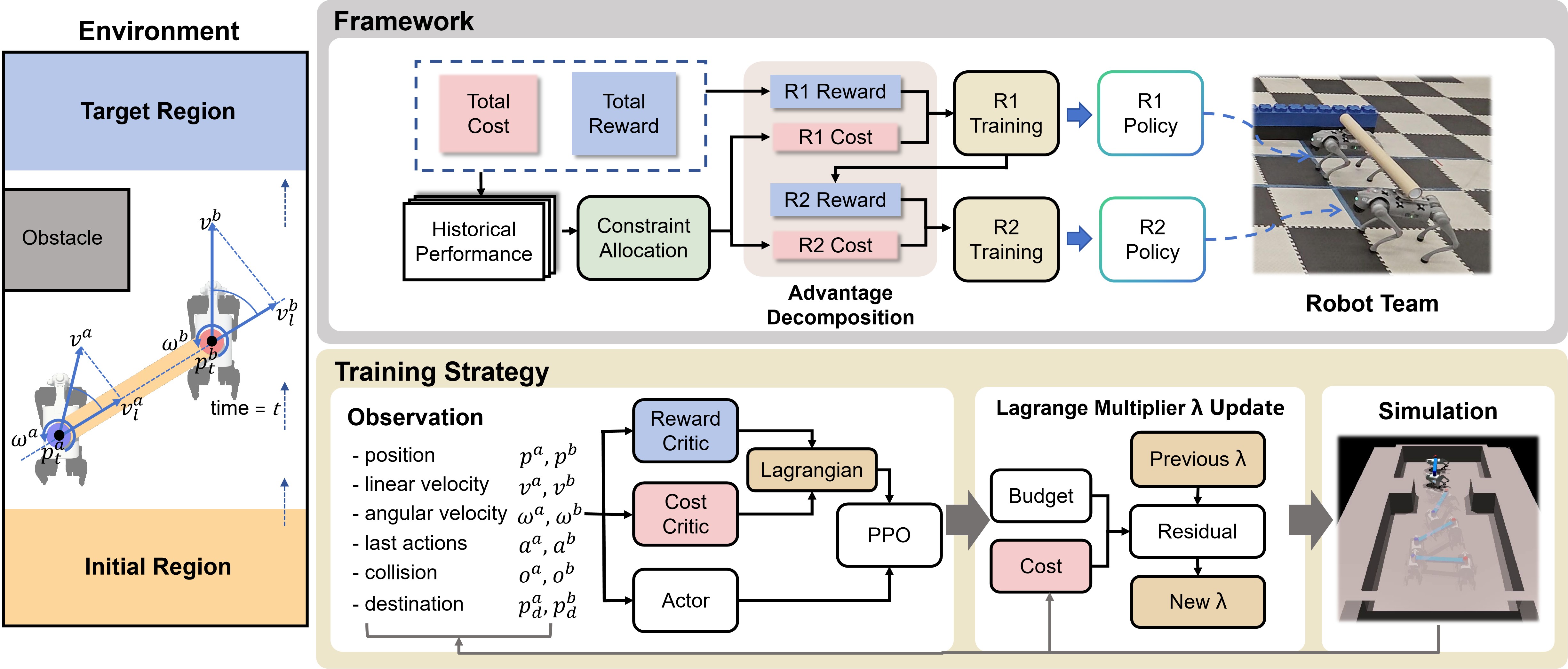}
    \caption{An overview of our method. 
    1) \textbf{Task Environment:} The robot team is required to collaboratively transport an object from the initial region to the target region while avoiding collisions.
    2) \textbf{Proposed Framework:} We formulate the task as a constrained Markov decision process.
For the total reward of the team, we adopt a safe RL approach to sequentially estimate the surrogate returns of R1 and R2, thereby determining the contribution of each member to the overall reward.
For the total cost of the team, we employ a constraint allocation method to assign individual cost budgets to each team member.
Robots R1 and R2 execute independent policies to accomplish the task in a fully distributed manner.
    3) \textbf{Training Strategy:} We use two separate critics to estimate the expected reward and cost, respectively. We adopt a Lagrangian approach in which the total advantage is computed by combining the reward and cost advantages. Then, the residual between the cost and the budget is used to update the Lagrange multiplier. Lastly, the updated policy is subsequently deployed in the simulation environment to collect data for the next iteration.
    }
    \label{fig:overview}
\end{figure*}

To address the above challenges, this paper proposes a RL framework (Fig. \ref{fig:overview}) for safe and high-performance decentralized collaborative transportation. 
Specifically, we formulate the transportation problem as a fully cooperative constrained Markov game, 
where task constraints are explicitly incorporated to guarantee the safety and performance of collaboration. 
A key challenge in this setting is that decentralized policy learning under shared constraints can lead to unstable learning dynamics and poor collaboration among robots.
To tackle this issue, we introduce a cost-advantage decomposition mechanism that enables each agent to estimate its contribution to the shared cost during policy learning, thereby improving learning stability under shared constraints. 
Building on this, we further propose a constraint allocation mechanism, which distributes team-level constraint budgets among agents and implicitly guides collaborative behaviors during learning. 
With these mechanisms, decentralized actor–critic learning can be carried out in a trust-region manner, ensuring monotonic improvement in both task performance and safety. 
Finally, a Lagrangian-based method is employed to efficiently enforce the allocated constraints during training.
The main contributions of this paper are as follows:

\begin{itemize}
    \item We propose a novel framework for decentralized collaborative transportation of dual quadruped robots in narrow environments. The proposed framework enhances inter-robot collaboration and enables the robots to safely transport a payload through highly constrained environments.
    \item We introduce a trust-region-based cost-advantage decomposition method into the RL framework to address safety challenges in collaborative transportation. This method enables the safety constraints to be fully enforced during policy learning, ensuring stable and safe policy updates within the RL process.
    \item We present a constraint allocation mechanism to improve performance in collaborative transportation tasks. By assigning different constraint budgets to individual robots, the proposed mechanism implicitly guides their coordinated behaviors and facilitates effective task-level collaboration.
    \item Simulations and real-world experiments demonstrate that our method outperforms existing approaches in decentralized dual-quadruped collaborative transportation task performance while ensuring safety.
\end{itemize}

%% file: 2ProblemStatement/problem_statement.tex



\section{PROBLEM STATEMENT}

\subsection{Constrained Markov Game}\label{section:A}
First of all, to formally describe the reinforcement learning process, the dual-quadruped collaborative transportation problem is modeled as a fully cooperative constrained Markov game, which is represented as a tuple:
\[
G = \langle \mathcal{N}, \mathcal{S}, \{\mathcal{A}_i\}_{i \in \mathcal{N}}, P, R, C, \gamma \rangle ,
\]
where:
\begin{itemize}
\item $\mathcal{N} = \{a, b\}$ is the set of robots and $\mathcal{S}$ is the state space;
\item $\mathcal{A}_i$ is the action space of robot $i$, with the action space $\boldsymbol{\mathcal{A}} = \mathcal{A}_a \times \mathcal{A}_b$;
\item $P: \mathcal{S} \times \boldsymbol{\mathcal{A}} \times \mathcal{S} \to [0,1]$ is the state transition probability;
\item $R: \mathcal{S} \times \boldsymbol{\mathcal{A}} \to \mathbb{R}$ and $C: \mathcal{S} \times \boldsymbol{\mathcal{A}} \to \mathbb{R}_+$ denote the joint reward function and the joint cost function of the system;
\item $\gamma \in [0,1)$ is the discount factor.
\end{itemize}
At each time step $t$, each robot $i$ observes a state $s_t \in \mathcal{S}$ and selects an action $a^i_t \in \mathcal{A}_i$, and $\boldsymbol{a}_t=\{a^a_t, a^b_t\}$. 
Given the action of all robots, the system receives a shared reward $R({s}_t,\boldsymbol{a}_t)$ and a shared cost $C({s}_t,\boldsymbol{a}_t)$.

For the reward part, 
we define the expected shared reward function $J_R(\boldsymbol{\pi})$: 
\begin{equation}
J_R(\boldsymbol{\pi}):=\mathbb{E}_{\boldsymbol{\pi}}\left[\sum_{t=0}^\infty\gamma^tR({s}_t,\boldsymbol{a}_t)\right],
\end{equation}
which is to be maximized under the policy $\boldsymbol{\pi}$. 
For the cost part, we define the shared cost function $J_C(\boldsymbol{\pi})$ under the policy $\boldsymbol{\pi}$:
\begin{equation}
J_C(\boldsymbol{\pi}):=\mathbb{E}_{\boldsymbol{\pi}}\left[\sum_{t=0}^\infty\gamma^tC({s}_t,\boldsymbol{a}_t)\right].
\end{equation}


Based on the above definitions, we describe the dual-quadruped collaborative transportation problem as:
\begin{equation}\label{original_problem}
\max_{\boldsymbol{\pi}}  J_R(\boldsymbol{\pi}) \quad
\text{s.t.} \quad J_C(\boldsymbol{\pi}) \leq u,
\end{equation}
where $u$ represents the threshold of the shared constraint of the system. 
Here, both the reward function $J_R(\boldsymbol{\pi})$ and cost function $J_C(\boldsymbol{\pi})$ are in the shared form. 
Hence, solving this problem in a decentralized system remains complicated. 

\subsection{Reward and Cost Design}
We adopt a hierarchical structure to solve the task. For the high-level layer, the observations of robot include: the destination position $\boldsymbol{p_d} \in \mathbb{R}^{2\times2}$, the robot's current position $\boldsymbol{p_t}\in \mathbb{R}^{2\times2}$, the robot’s last position $\boldsymbol{p_{t-1}}\in \mathbb{R}^{2\times2}$, linear velocity $\boldsymbol{v}\in \mathbb{R}^{2\times2}$, angular velocity $\boldsymbol{\omega}\in \mathbb{R}^{1\times2}$, 
yaw angle $\boldsymbol{\theta} \in \mathbb{R}^{1\times2}$, and the feedback from collision detection points (4 points for each robot) around the robot $\boldsymbol{o} \in \mathbb{R}^{4\times2}$. 
The actions of robot include: the target linear velocity $\boldsymbol{a_v} \in \mathbb{R}^{2\times2}$ and the target angular velocity $\boldsymbol{a_\omega}\in \mathbb{R}^{1\times2}$ around the $z$-axis. 
We define an indicator function $I(\cdot)$ which returns $1$ if the condition inside the parentheses is true, and $0$ otherwise.
For any time step $t$, we define the following reward and cost functions.

\textbf{Move-Forward Reward}: This function rewards the system for moving toward the destination:
\begin{equation}
    r^t_f\left(\boldsymbol{p_t}, \boldsymbol{p_{t-1}}, \boldsymbol{p_d}\right)=\omega_f\cdot I\left(\|\boldsymbol{p_t}-\boldsymbol{p_d}\|_2<\|\boldsymbol{p_{t-1}}-\boldsymbol{p_d}\|_2\right)
\end{equation}

\textbf{Destination-Reached Reward}: This function provides a reward when the system reaches the destination:
\begin{equation}
    r^t_d\left(\boldsymbol{p_t}, \boldsymbol{p_d}\right)=\omega_d\cdot I\left(\|\boldsymbol{p_t}-\boldsymbol{p_d}\|_2<b\right)
\end{equation}

\textbf{Collaborative Task Cost}: This function provides a cost when linear velocity command of robots along the connecting link $a^a_l, a^b_l$ are in opposite directions:
\begin{equation}
    c^t_m\left({a^a_l}, {a^b_l}\right)=\omega_m\cdot I\left({a^a_l}\times{a^b_l}<0\right)
\end{equation}

\textbf{Safety Task Cost}: This function provides a cost when collisions occur, with $h$ denoting the number of collision detection points:
\begin{equation}
    c^t_c\left(\boldsymbol{o}\right)=\omega_c\sum_{i=1}^h I\left({o_i}>0\right)
\end{equation}

For the lower-level controller, we  apply the pretrained model \cite{margolis2023walk}, which controls the velocity of robots. 
In this work, for each robot, the high-level layer generates velocity commands to drive the robot toward the goal as efficiently as possible ($r^t_f$ and $r^t_d$), 
while coordinating the velocities between the two robots ($c^t_m$) and avoiding collisions with obstacles ($c^t_c$).
The low-level controller then tracks these velocity commands and outputs the corresponding joint positions.

%% file: 3ProblemSolving/problem_solving.tex
\section{PROBLEM SOLVING}

\subsection{Advantage Decomposition}\label{section:cost}
To address the collaborative transportation problem \eqref{original_problem} under safety constraints, we propose a trust-region based collaborative transportation framework.
We introduce a generic scalar signal $X \in \{R, C\}$, where $X$ denotes either the reward $R$ or the cost $C$. 
For any such signal $X$, we define the unified state-action value function and state-value function under policy $\boldsymbol{\pi}$ as follows:
\begin{align}
Q_{X,\boldsymbol{\pi}}(s, \boldsymbol{a}) &:= \mathbb{E}_{\boldsymbol{\pi}} \left[ \sum_{t=0}^\infty \gamma^t X(s_t, \boldsymbol{a_t}) \,\middle|\, s_0 = s, \boldsymbol{a_0} = a \right],\nonumber\\
V_{X,\boldsymbol{\pi}}(s) &:= \mathbb{E}_{\boldsymbol{\pi}} \left[ \sum_{t=0}^\infty \gamma^t X(s_t, a_{t}) \,\middle|\, s_0 = s \right].
\end{align}
We use a randomly sampled update order of each robot to promote convergence toward a Nash equilibrium.
Let the indices $i_1$ and $i_2$ denote the first and second robots to be updated in a given iteration (i.e., the update order, not fixed robot identities), 
$a^{i_1}$ and $a^{i_2}$ denote the corresponding actions, $\pi^{i_1}$ and $\pi^{i_2}$ denote the corresponding policies. 
The marginal values conditioned on the action of the first and second robots are defined as:
\begin{align}
Q_{X,\boldsymbol{\pi}}^{i_1}(s, a^{i_1}) &:= \mathbb{E}_{a^{i_2} \sim \pi^{i_2}(\cdot \mid s)} \left[ Q_{X,\boldsymbol{\pi}}(s, \boldsymbol{a}) \right], \\
Q_{X,\boldsymbol{\pi}}^{i_2}(s, a^{i_2}) &:= \mathbb{E}_{a^{i_1} \sim \pi^{i_1}(\cdot \mid s)} \left[ Q_{X,\boldsymbol{\pi}}(s, \boldsymbol{a}) \right].
\end{align}
In addition, the advantage functions of the first and second updated robots are defined as:
\begin{align}
A_{X,\boldsymbol{\pi}}^{i_1}(s, a^{i_1}) &:= Q_{X,\boldsymbol{\pi}}^{i_1}(s, a^{i_1}) - V_{X,\boldsymbol{\pi}}(s), \\
A_{X,\boldsymbol{\pi}}^{i_{1,2}}(s, a^{i_1}, a^{i_2}) &:= Q_{X,\boldsymbol{\pi}}(s, \boldsymbol{a}) - Q_{X,\boldsymbol{\pi}}^{i_1}(s, a^{i_1}).
\end{align}
Let $\hat{\pi}$ and $\bar{\pi}$ denote candidate and updated policy, respectively. 
The surrogate returns of the first and second updated robots are defined as:
\begin{align}
L_{X,\boldsymbol{\pi}}^{i_1}(\hat{\pi}^{i_1}) &:= \mathbb{E}_{a^{i_1} \sim \hat{\pi}^{i_1}} \left[ A_{X,\boldsymbol{\pi}}^{i_1}(s, a^{i_1}) \right], \\
L_{X,\boldsymbol{\pi}}^{i_{1,2}}(\bar{\pi}^{i_1}, \hat{\pi}^{i_2}) &:= \mathbb{E}_{a^{i_1} \sim \bar{\pi}^{i_1},\, a^{i_2} \sim \hat{\pi}^{i_2}} \left[ A_{X,\boldsymbol{\pi}}^{i_{1,2}}(s, a^{i_1}, a^{i_2}) \right].
\end{align}
Based on the above definitions, we investigate the properties that support our framework. 
For any $X \in \{R, C\}$, the system advantage function $A_{X,\boldsymbol{\pi}}(s, \boldsymbol{a})$ can be decomposed into the sum of the advantage functions of the two robots:
\begin{align}
&A_{X,\boldsymbol{\pi}}(s, \boldsymbol{a})\nonumber \\
           &= Q_{X,\boldsymbol{\pi}}(s, \boldsymbol{a}) - V_{X,\boldsymbol{\pi}}(s)\nonumber \\
           &= \left[ Q_{X,\boldsymbol{\pi}}^{i_1}(s, a^{i_1}) - V_{X,\boldsymbol{\pi}}(s) \right] + \left[ Q_{X,\boldsymbol{\pi}}(s, \boldsymbol{a}) - Q_{X,\boldsymbol{\pi}}^{i_1}(s, a^{i_1}) \right]\nonumber \\
           &= A_{X,\boldsymbol{\pi}}^{i_1}(s, a^{i_1}) + A_{X,\boldsymbol{\pi}}^{i_{1,2}}(s, a^{i_1}, a^{i_2}).
\end{align}
For the reward term $R$, applying the trust-region analysis to the generic signal $X$ yields a lower bound of the updated reward function $J_R(\bar{\pi})$ 
(proved by Kuba et al. \cite{kuba2022trust}):
\begin{gather}
J_R(\bar{\pi}) \geq J_R(\pi) 
+ L_{R,\boldsymbol{\pi}}^{i_1}(\bar{\pi}^{i_1}) - \nu_R D_{\mathrm{KL}}^{\max}(\pi^{i_1}, \bar{\pi}^{i_1}) \nonumber \\
\phantom{J_R(\bar{\pi}) \geq}
+ L_{R,\boldsymbol{\pi}}^{i_{1,2}}(\bar{\pi}^{i_1}, \bar{\pi}^{i_2}) - \nu_R D_{\mathrm{KL}}^{\max}(\pi^{i_2}, \bar{\pi}^{i_2}), 
\label{prop_reward}
\end{gather}
where $D_{\mathrm{KL}}^{\max}(\pi^{i}, \hat{\pi}^{i}) := \max_s D_{\mathrm{KL}}(\pi^{i}(\cdot \mid s), \hat{\pi}^{i}(\cdot \mid s))$ and $\nu_R = \frac{4\gamma \max_{s,a} |A_{R,\boldsymbol{\pi}}(s, \boldsymbol{a})|}{(1 - \gamma)^2}$. 
For the cost term $C$, the higher bound of the updated cost function $J_C(\bar{\pi})$ exists:
\begin{gather}
J_C(\bar{\pi}) \leq J_C(\pi) 
+ L_{C,\boldsymbol{\pi}}^{i_1}(\pi^{i_1}) + \nu_C D_{\mathrm{KL}}^{\max}(\pi^{i_1}, \bar{\pi}^{i_1}) \nonumber \\
\phantom{J_C(\bar{\pi}) \leq}
+ L_{C,\boldsymbol{\pi}}^{i_{1,2}}(\pi^{i_1}, \bar{\pi}^{i_2}) + \nu_C D_{\mathrm{KL}}^{\max}(\pi^{i_2}, \bar{\pi}^{i_2}), 
\label{prop_cost}
\end{gather}
where $\nu_C = \frac{4\gamma \max_{s,a} |A_{C,\boldsymbol{\pi}}(s, \boldsymbol{a})|}{(1 - \gamma)^2}$. 
Based on the properties above, the collaborative transportation problem \eqref{original_problem} can be reformulated as follows:
\begin{equation}\label{problem2}\begin{aligned}
\max_{\pi^{i_1},\pi^{i_2}}\quad &L_{R,\boldsymbol{\pi}}^{i_1}(\bar{\pi}^{i_1}) + L_{R,\boldsymbol{\pi}}^{i_{1,2}}(\bar{\pi}^{i_1}, \bar{\pi}^{i_2}),\\
\text{s.t.}\quad &L_{C,\boldsymbol{\pi}}^{i_1}(\pi^{i_1}) + L_{C,\boldsymbol{\pi}}^{i_{1,2}}(\pi^{i_1}, \bar{\pi}^{i_2}) \leq d,\\
\quad & D_{\mathrm{KL}}^{\max}(\pi^{i_1}, \bar{\pi}^{i_1}) \leq \delta,\  D_{\mathrm{KL}}^{\max}(\pi^{i_2}, \bar{\pi}^{i_2}) \leq \delta,
\end{aligned}\end{equation}
where $d = u - J_C(\boldsymbol{\pi})$ represents the cost threshold of the system. 
By \eqref{prop_reward}, as long as the KL divergences $D_{\mathrm{KL}}^{\max}(\cdot)$ are small enough, the max term ensures that each updated policy achieves a maximally monotonic performance improvement. 
In addition, by \eqref{prop_cost}, the safety joint constraint $J_C(\boldsymbol{\pi}) \leq u$ is simplified into a summation form. 
However, this reformulated problem still retains a joint structure, remaining difficult to solve the policies of individual robots independently. 
Therefore, in the next section, we propose the constraint allocation method to address the joint structure.

\subsection{Constraint Allocation} \label{section:constraint}
In this section, 
we propose the constraint allocation method designed to address the difficulty of joint optimization and to enhance team collaboration. 
For the two-robot system, we label the two robots as $a$ and $b$. 
It is important to note that $a$ and $b$ are different from $i_1$ and $i_2$: 
the former denotes fixed robot identities, whereas the latter indicates the update order of robots in an iteration, 
since the update order is randomized each time. 

Denote $c_a$ and $c_b$ as the constraint values allocated to robots $a$ and $b$, respectively, 
the joint constraint in~\eqref{problem2} can be transformed into:
\begin{gather}
L_{C,\boldsymbol{\pi}}^{a}(\pi^{a}) \leq c_a \quad \text{for robot}\ a,\nonumber\\
L_{C,\boldsymbol{\pi}}^{b}(\pi^{b}) \leq c_b \quad \text{for robot}\ b,\\
\text{with}\quad c_a+c_b=d.\nonumber
\end{gather}
Here, $L_{C,\boldsymbol{\pi}}^a(\pi_a)$ and $L_{C,\boldsymbol{\pi}}^b(\pi_b)$ denote the surrogate cost returns for robots $a$ and $b$, respectively. 
Specifically, if robot $a$ is updated first ($i_1 = a$), then $L_{C,\boldsymbol{\pi}}^a(\pi_a) := L_{C,\boldsymbol{\pi}}^{i_1}(\pi_a)$; 
if updated second ($i_2 = a$), then $L_{C,\boldsymbol{\pi}}^a(\pi_a) := L_{C,\boldsymbol{\pi}}^{i_1,i_2}(\bar{\pi_b}, \pi_a)$. 
The same rule applies to robot $b$.

Importantly, the choice of constraint values $(c_a,c_b)$ influences the collaborative behavior. 
For instance, allocating a larger $c_a$ to robot $a$ permits more aggressive policy updates, 
potentially encouraging it to assume a leading role in navigation. 
Conversely, a tighter budget induces more conservative behavior. 
This observation motivates us to optimize the allocation itself to maximize overall team performance.

Let $\pi(c_a,c_b)$ denote the joint policy obtained under allocation $(c_a,c_b)$.
We define the performance objective as a weighted combination of task reward and system cost:
\begin{equation}
F(c_a,c_b) := w_1 J_R(\pi(c_a,c_b)) - w_2 J_C(\pi(c_a,c_b)),
\end{equation}
where $w_1$ and $w_2$ are non-negative user-defined weights that balance the task reward and system cost. 
Since the constraint $c_a + c_b = d$ holds, once $c_a$ is obtained, $c_b$ can be directly determined. 
Thus, we formulate the allocation problem as the following optimization problem:
\begin{equation}
c_a^* = \arg\max_{c_a} F(c_a)\quad \text{and}\quad c_b^*=d-c_a^*.
\end{equation}
We simplify the notation by writing $F(c_a,c_b)$ as $F(c_a)$, as $c_b$ is uniquely determined by $c_a$. 
However, $F(c_a)$ is an unknown, non-convex, and potentially noisy function. 
To address this, we model $F(c_a)$ as a black-box function and employ Bayesian optimization for efficient search. 
Specifically, we place a Gaussian process $\mathcal{G}$ prior over $F(c_a)$:
\begin{equation}
F(c_a) \sim \mathcal{G}(m(c_a), k(c_a, c_a')),
\end{equation}
where $m(c_a)$ and $k(c_a, c_a')$ denote the mean and covariance functions of the Gaussian process prior over $F(c_a)$. 
In addition, we maintain a sliding window of the most recent $W$ observations to adapt to the non-stationarities induced by policy learning.
Let $y$ represent the observation of $F(c_a)$, and let $\mathcal{I}_t := [t - W + 1,\, t]$ denote the sampling interval. 
The $t$-th historical observation is defined as:
\begin{equation}
\mathcal{D}_t := \left\{ (c_a^{(k)}, y^{(k)}) \right\}_{k\in \mathcal{I}}.
\end{equation}
For any new candidate point $c_a'$, 
the posterior distribution of its corresponding objective value $y$ is Gaussian:
\begin{equation}
\phantom{\text{where}} \quad 
y \mid \mathcal{D}_t, c_a' \sim \mathcal{N}(\mu(c_a'), \sigma^2(c_a')), 
\end{equation}
where $\mu_t(c_a') = \mathbf{k}'^\top \left( \mathbf{K} + \sigma_n^2 \mathbf{I} \right)^{-1} \mathbf{y}_t$ 
and $\sigma_t^2(c_a') = k(c_a', c_a') - \mathbf{k}'^\top \left( \mathbf{K} + \sigma_n^2 \mathbf{I} \right)^{-1} \mathbf{k}'$ 
denote the posterior mean and variance at candidate point $c_a'$, 
$\mathbf{K} \in \mathbb{R}^{W \times W}$ is the kernel matrix with elements $K_{ij} = k(c_a^{(i)}, c_a^{(j)})$, 
$\mathbf{k}' \in \mathbb{R}^W$ is the vector of cross-covariances with elements $k(c_a', c_a^{(i)})$, 
$\mathbf{y}_t = \left[ y^{(t-W+1)}, \dots, y^{(t)} \right]^\top$,
and $\sigma_n^2$ denotes the observation noise variance.

To obtain the best performance under the current observations, we adopt the Expected Improvement (EI) as the acquisition function:
\begin{equation}
\mathrm{EI}(c_a) = \mathbb{E}\left[ \max\{0, y - y_{\text{best}}\} \right],
\end{equation}
where $y_{\text{best}} = \max_{k\in \mathcal{I}} y^{(k)}$. 
Then, we denote $\Phi(\cdot)$ and $\phi(\cdot)$ as the cumulative distribution function and the probability density function of the standard normal distribution, respectively. 
When $\sigma(c_a) = 0$, we have $\mathrm{EI}(c_a) = 0$; when $\sigma(c_a) > 0$, $\mathrm{EI}(c_a)$ admits the following closed-form expression:
\begin{equation}
\mathrm{EI}(c_a) =
\left( \mu(c_a) - y_{\text{best}} \right) {\Phi}\left( z \right) + \sigma(c_a) \phi\left( z \right), 
\end{equation}
where $\quad z=\dfrac{\mu(c_a) - y_{\text{best}}}{\sigma(c_a)}$. The optimal allocation $(c_a^*,c_b^*)$ is then obtained by:
\begin{align}
    c_a^* &= \arg\max_{c_a} \mathrm{EI}(c_a),\\
    c_b^* &= d - c_a^*.
\end{align}
At this point, we obtain the constraint allocation $(c_a^*,c_b^*)$ that improves the team performance. 


\subsection{Training Strategy}\label{section:lag}

After constraint allocation, robot $a$ and $b$ need to solve their constrained individual optimization problems during their update turn. 
Specifically, for robot $a$, its optimization problem can be formulated as:
\begin{equation}
\begin{split}
&\boldsymbol{\bar{\pi}}^{a} = \arg\max_{\boldsymbol{\bar{\pi}}^{a}} \left[ \mathcal{L}_{R,\boldsymbol{\pi}}^{a}(\pi^{a}) \right],\\
\text{s.t.} \quad &\mathcal{L}_{C,\boldsymbol{\pi}}^{a}(\boldsymbol{\pi}^{a}) \leq c_a, \\
& D_{\mathrm{KL}}^{\max}(\pi^{a}, \bar{\pi}^{a}) \leq \delta.
\end{split}
\end{equation}
To solve this constrained optimization problem, we employ Lagrangian relaxation. 
Let $\lambda_a$ denote a non-negative Lagrange multiplier. 
The updated policy of robot $a$ can be obtained by solving the following max-min problem:
\begin{equation}\label{eq:lag}
\begin{split}
    \max_{\bar{\pi}^{a}} \min_{\lambda_a \geq 0}\quad& 
    \mathcal{L}_{R,\boldsymbol{\pi}}^{a}(\boldsymbol{\pi}^{a}) 
    - \lambda_a \left[ \mathcal{L}_{C,\boldsymbol{\pi}}^{a}(\pi_{a}) - c_a \right], \\
    &\text{s.t.}\quad D_{\mathrm{KL}}^{\max}(\pi^{a}, \bar{\pi}^{a}) \leq \delta.
\end{split}
\end{equation}
The Lagrange multiplier $\lambda_a$ is then updated based on the residual, which is the difference between the surrogate cost $\mathcal{L}_{C,\boldsymbol{\pi}}^{a}(\boldsymbol{\pi}^{a})$ and the budget $c_a$. 
Denote $\lambda_a^i$ as the value of $\lambda_a$ at the $i$-th iteration, and $\alpha$ as a constant coefficient. The update of $\lambda_a$ can be expressed as:
\begin{equation}\label{eq:lamda_update}
\lambda^{i+1}_a = \lambda^i_a + \alpha(\mathcal{L}_{C,\boldsymbol{\pi}}^{a}(\pi_{a}) - c_a)
\end{equation}
%
By~\eqref{eq:lag}, robot $a$ improves its policy under a dual mechanism: 
The PPO clipping mechanism ensures moderate policy updates, while the Lagrange multiplier $\lambda_a$ penalizes cost advantage to satisfy constraints, 
leading to safe and efficient policy improvement.
The same optimization process applies to robot $b$.

%% file: 4Simulation/simulation.tex
\begin{table}[H]
\centering
\caption{Comparison of Baseline Methods}
\label{tab:baseline}
\begin{tabular}{lcccc}
\hline
\textbf{Method} 
& \textbf{\makecell{Explicit\\Constraint}} 
& \textbf{\makecell{Constraint\\Allocation}} 
& \textbf{\makecell{Independent\\Policies}} 
& \textbf{\makecell{Lagrangian}} \\
\hline
MAPPO  & \xmark & -- & \xmark & -- \\
HAPPO  & \xmark & -- & \cmark & -- \\
UCA    & \cmark & \xmark & \cmark & \cmark \\
MACPO  & \cmark & \cmark & \xmark & \cmark \\
\textbf{Ours}   & \cmark & \cmark & \cmark & \cmark \\
\hline
\end{tabular}
\end{table}

\begin{figure}[H]
  \centering
  \includegraphics[width=\columnwidth]{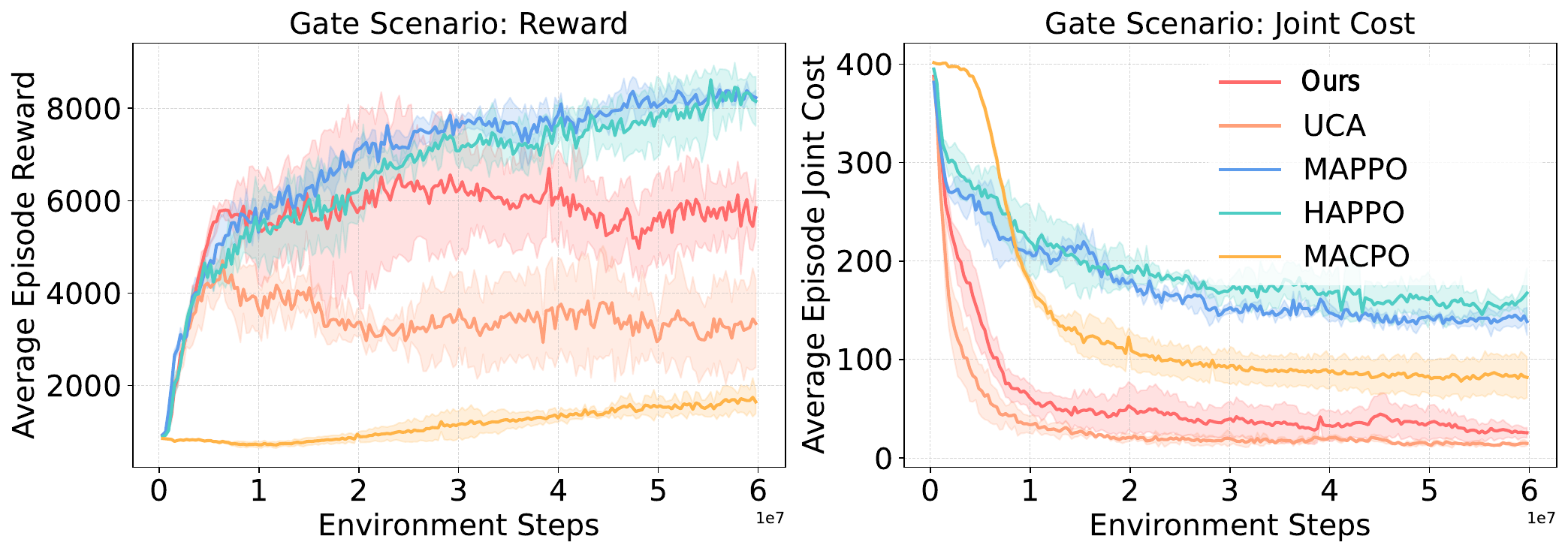}
  \caption{Average episode reward (left) and cost (right) during training. Our method (red curve) achieves high reward while maintaining a low cost, suggesting that it effectively balances safety and efficiency.}
  \label{fig:sim_r_c}
\end{figure}

\begin{table}[H]
\centering
\captionsetup{justification=centering}
\caption{Safety and Collaboration Efficiency of Different Methods in the Gate Scenario over 30 Trials\\
($\text{Trajectory Straightness} = \frac{\text{Displacement Along  Goal Direction}}{\text{Trajectory Length}}$)
}
\label{tab:safety_gate}

\begin{tabular}{lcccc}
\toprule
\textbf{Method}
& \textbf{\makecell{Collision\\Rate [\%]}}
& \textbf{\makecell{Arrival\\Rate [\%]}}
& \textbf{\makecell{Trajectory\\Straightness}}
& \textbf{\makecell{Time\\Consumption [s]}} \\
\midrule
MAPPO          & 90.0         & 10.0           & --            & --            \\
HAPPO          & 83.3         & 16.7           & --            & --            \\
UCA            & 6.7          & 53.3           & 0.68          & 26.8          \\
MACPO          & 0.0          & 6.7            & 0.74          & 32.0          \\
\textbf{Ours}  & \textbf{0.0} & \textbf{100.0} & \textbf{0.95} & \textbf{18.3} \\
\bottomrule
\end{tabular}

\end{table}







\FloatBarrier
\section{SIMULATION EXPERIMENTS}

\begin{figure}[t]
    \centering
    \begin{minipage}[b]{0.4\columnwidth}
        \centering
        \includegraphics[width=\linewidth]{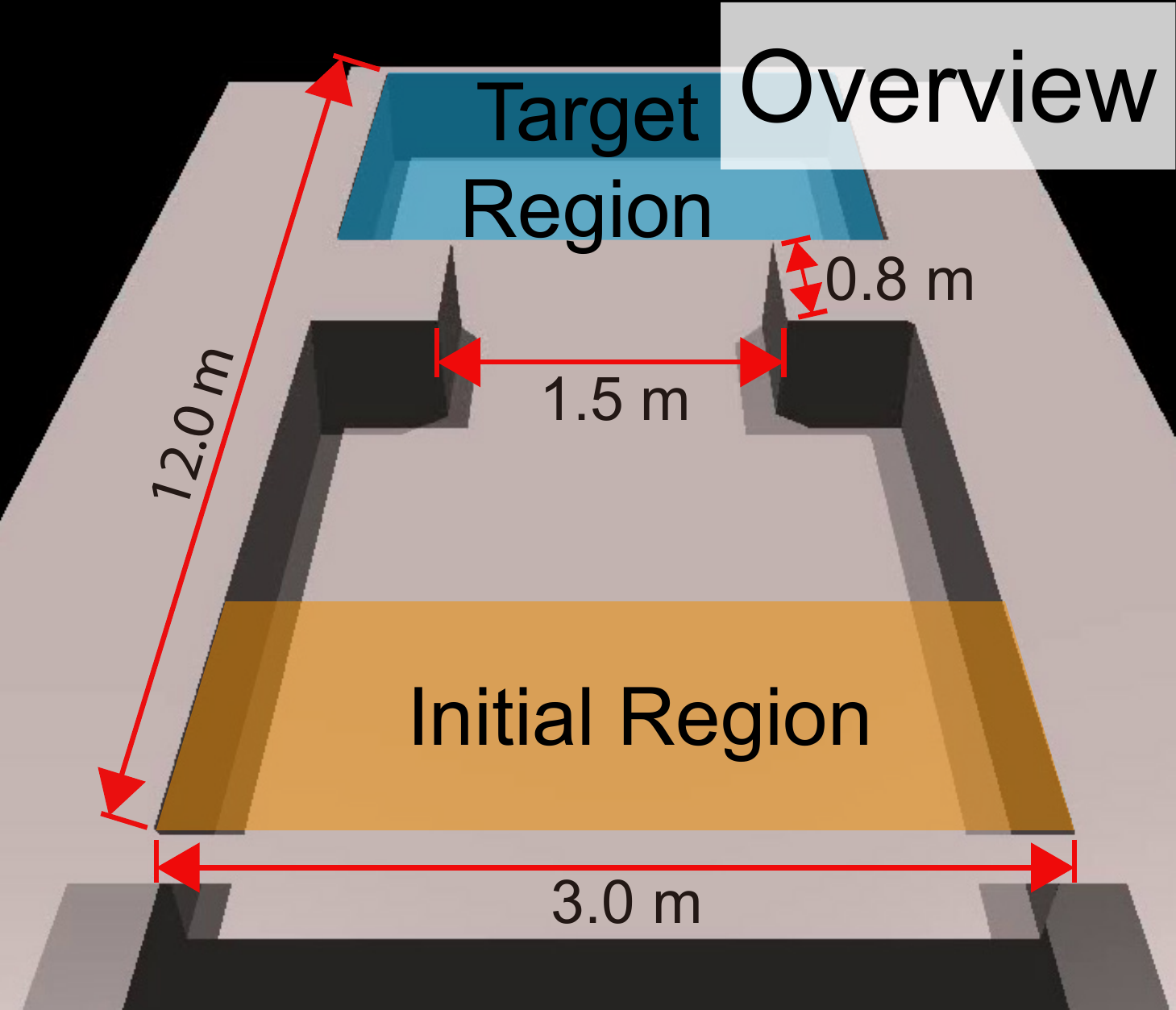}
        \subcaption{}
    \end{minipage}
    \begin{minipage}[b]{0.4\columnwidth}
        \centering
        \includegraphics[width=\linewidth]{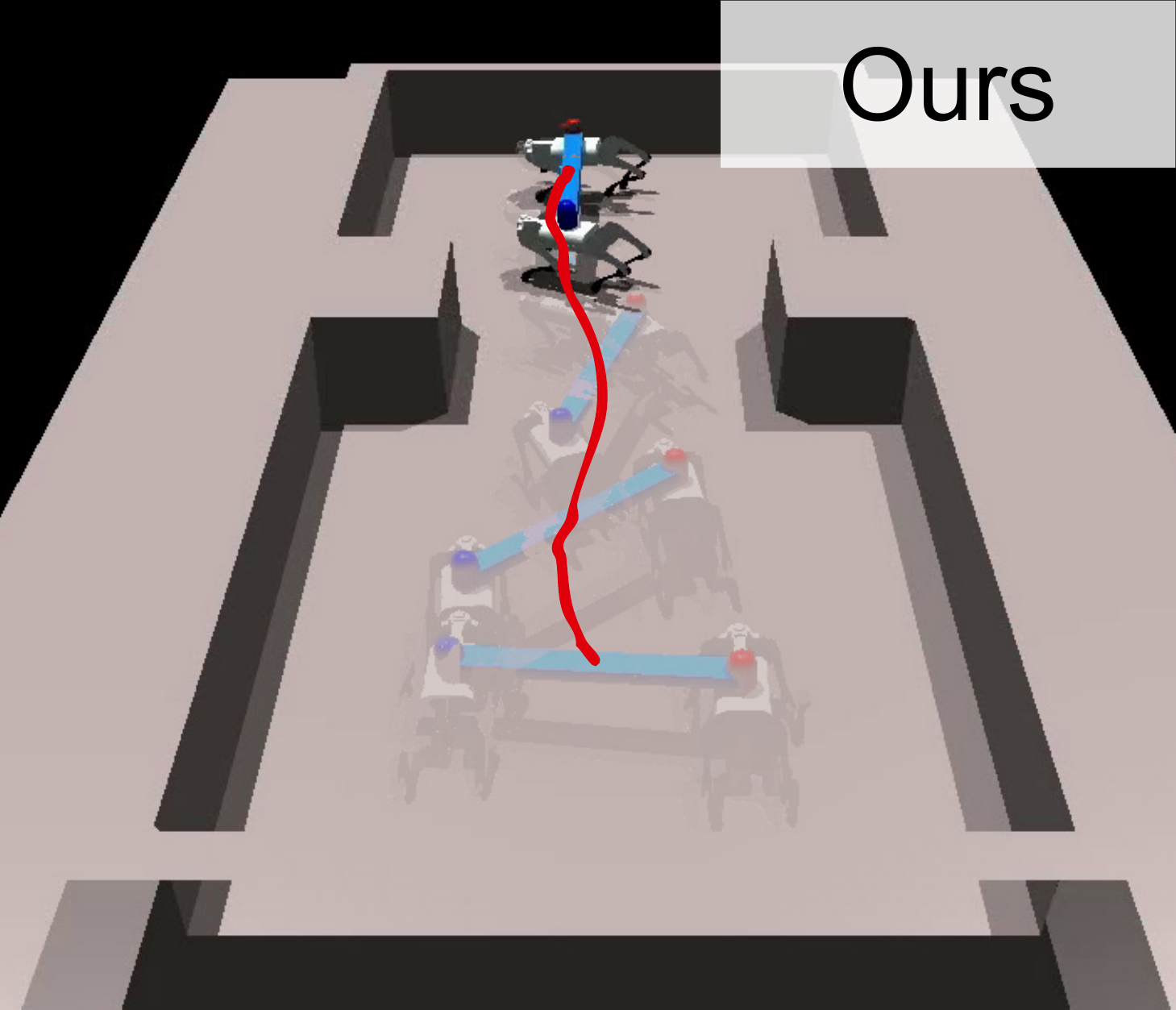}
        \subcaption{}
    \end{minipage}

    \vspace{0.3em} 

    \centering
    \begin{minipage}[b]{0.4\columnwidth}
        \centering
        \includegraphics[width=\linewidth]{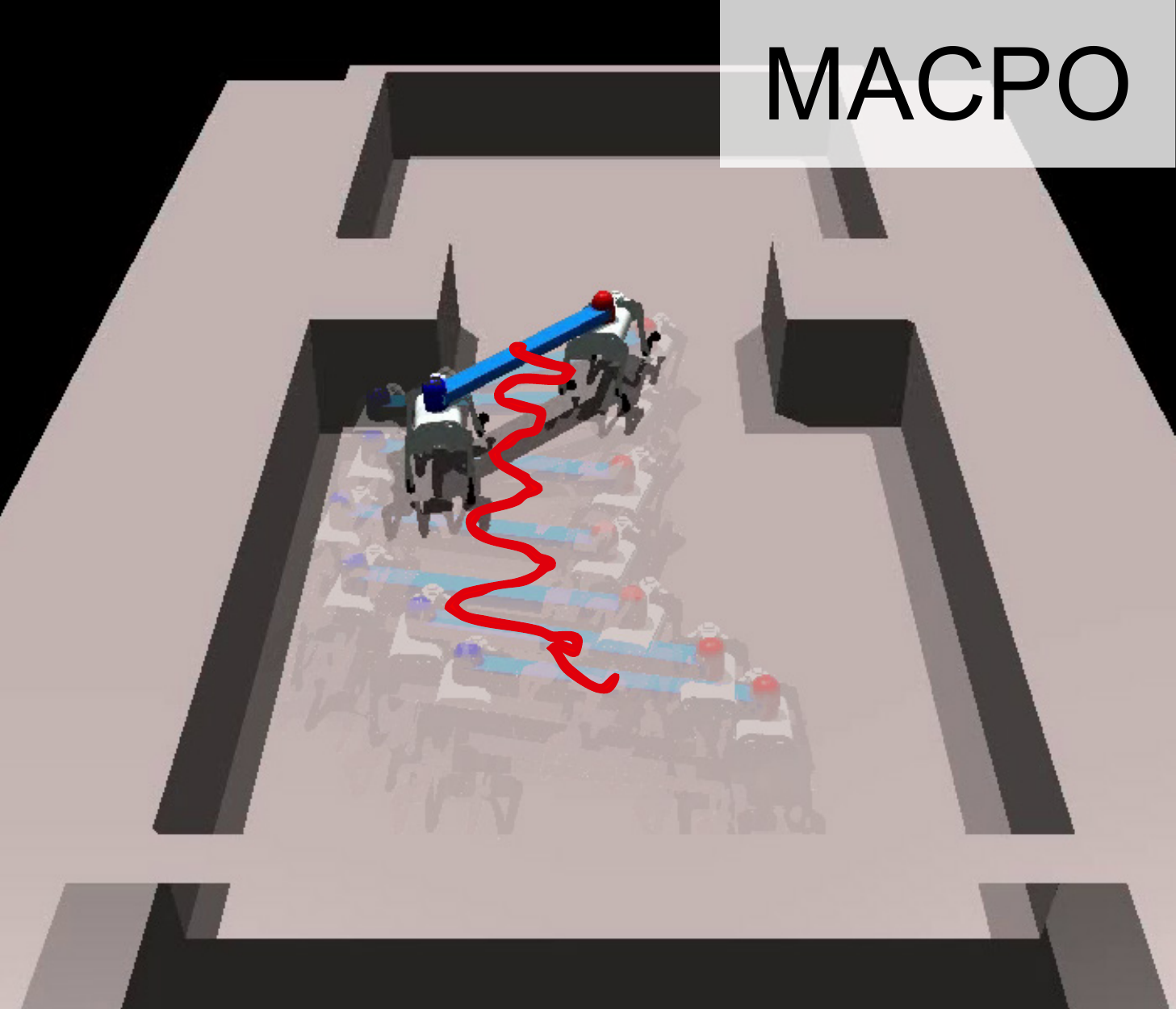}
        \subcaption{}
    \end{minipage}
    \begin{minipage}[b]{0.4\columnwidth}
        \centering
        \includegraphics[width=\linewidth]{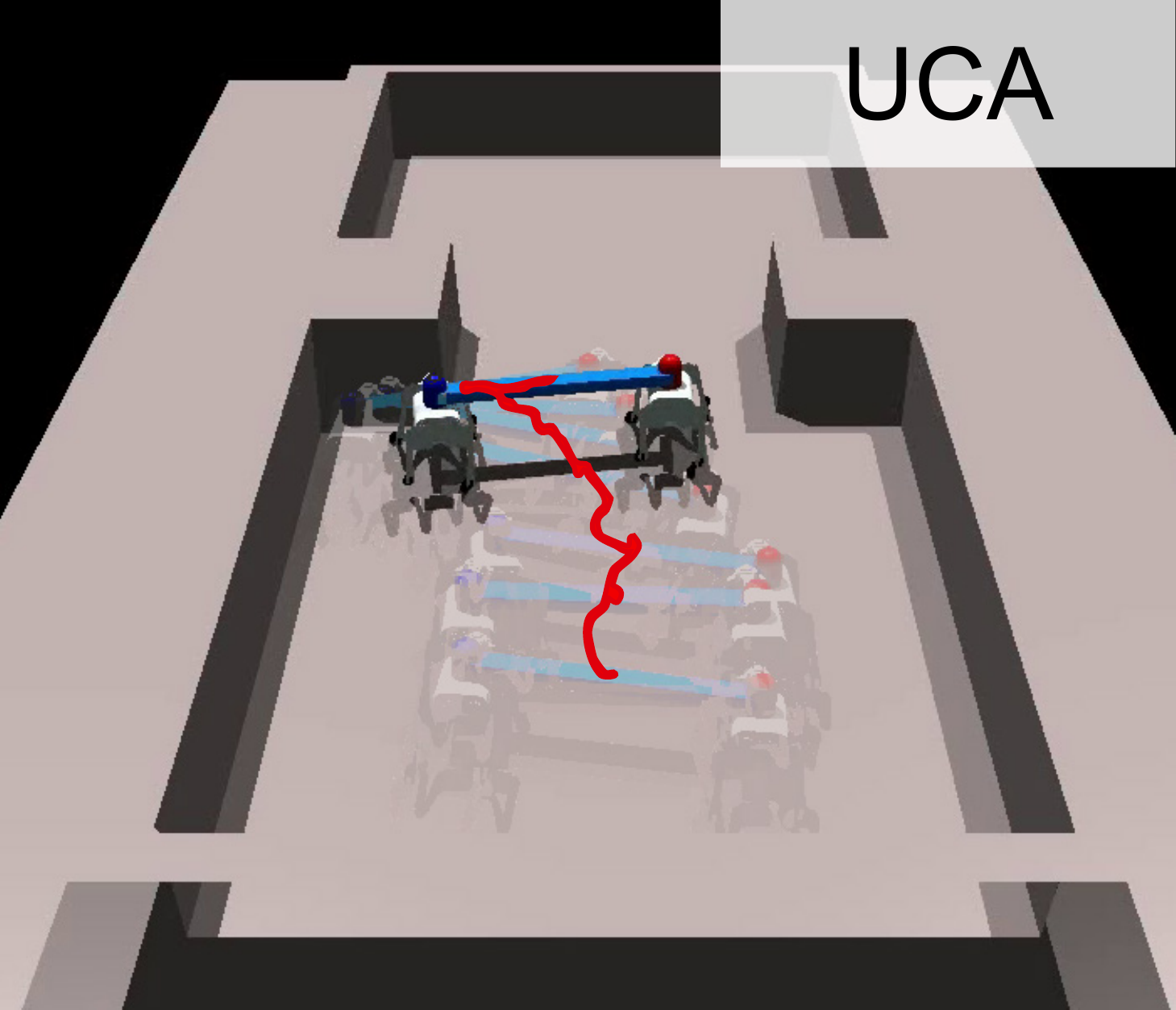}
        \subcaption{}
    \end{minipage}

    \vspace{0.3em} 

    \centering
    \begin{minipage}[b]{0.4\columnwidth}
        \centering
        \includegraphics[width=\linewidth]{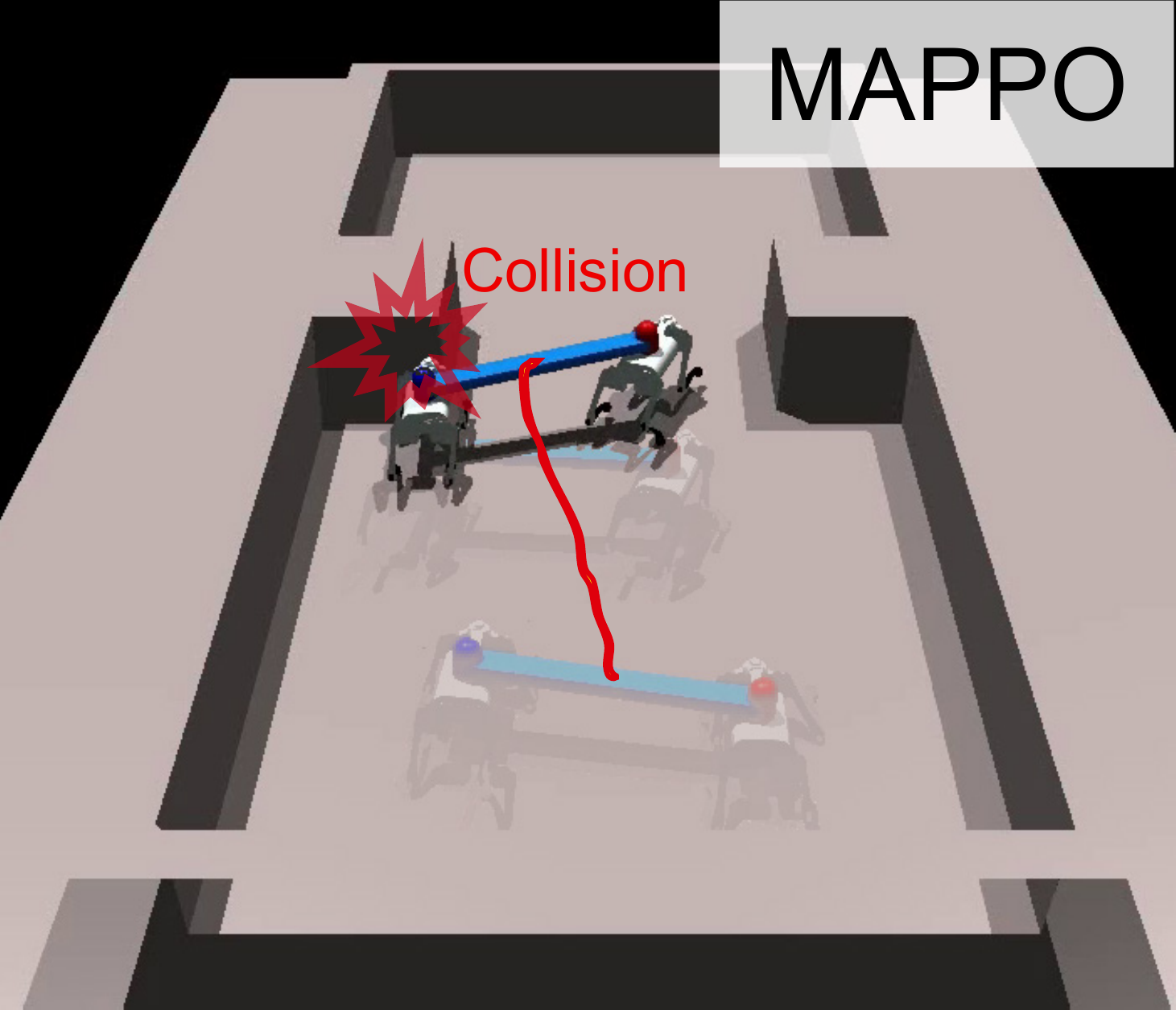}
        \subcaption{}
    \end{minipage}
    \begin{minipage}[b]{0.4\columnwidth}
        \centering
        \includegraphics[width=\linewidth]{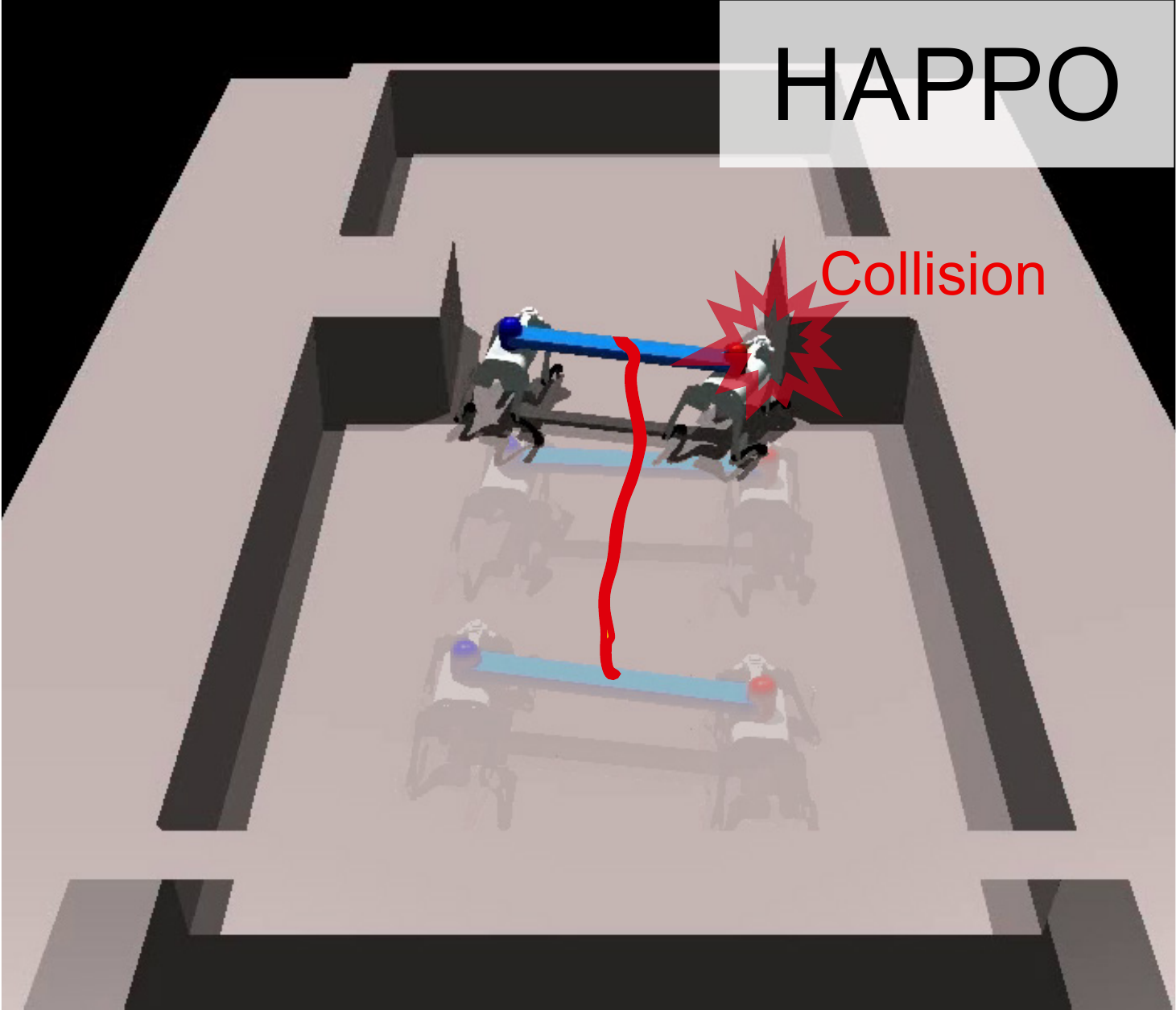}
        \subcaption{}
    \end{minipage}

    \caption{(a): Parameters of the gate scenario. The robot team is required to traverse the environment from the initial region to the target region without collisions.
     (b): The robot team successfully reaches the target region within the specified time horizon. The red line denotes the trajectory of the payload.
     (c) \& (d): The robot team fails to reach the target region within the specified time horizon. The payload trajectory is distorted.
  (e) \& (f): The robots collide with obstacles during navigation.
    }
    \label{fig:sim_gate}
\end{figure}

In this section, we validate the effectiveness of our method through simulation experiments, especially focusing on the safety and performance of collaboration. 
Specifically, we construct a narrow-gate scenario (Fig.~\ref{fig:sim_gate}) where the gate has a width of $1.5$~m and a depth of $0.8$~m. 
In this scenario, robots are required to collaboratively adjust their formation to pass through the narrow gate and reach the target location, 
while avoiding collisions throughout the process. 
The maximum length of each episode is $200$ steps. 
Both training and simulation experiments are conducted on a computing platform equipped with an Intel Core i9 processor and an NVIDIA GeForce RTX 4080 GPU, 
and the simulation environment is built using Isaac Gym.








\begin{figure}[t]
    \centering
    \begin{minipage}[b]{0.48\columnwidth}
        \centering
        \includegraphics[width=\linewidth]{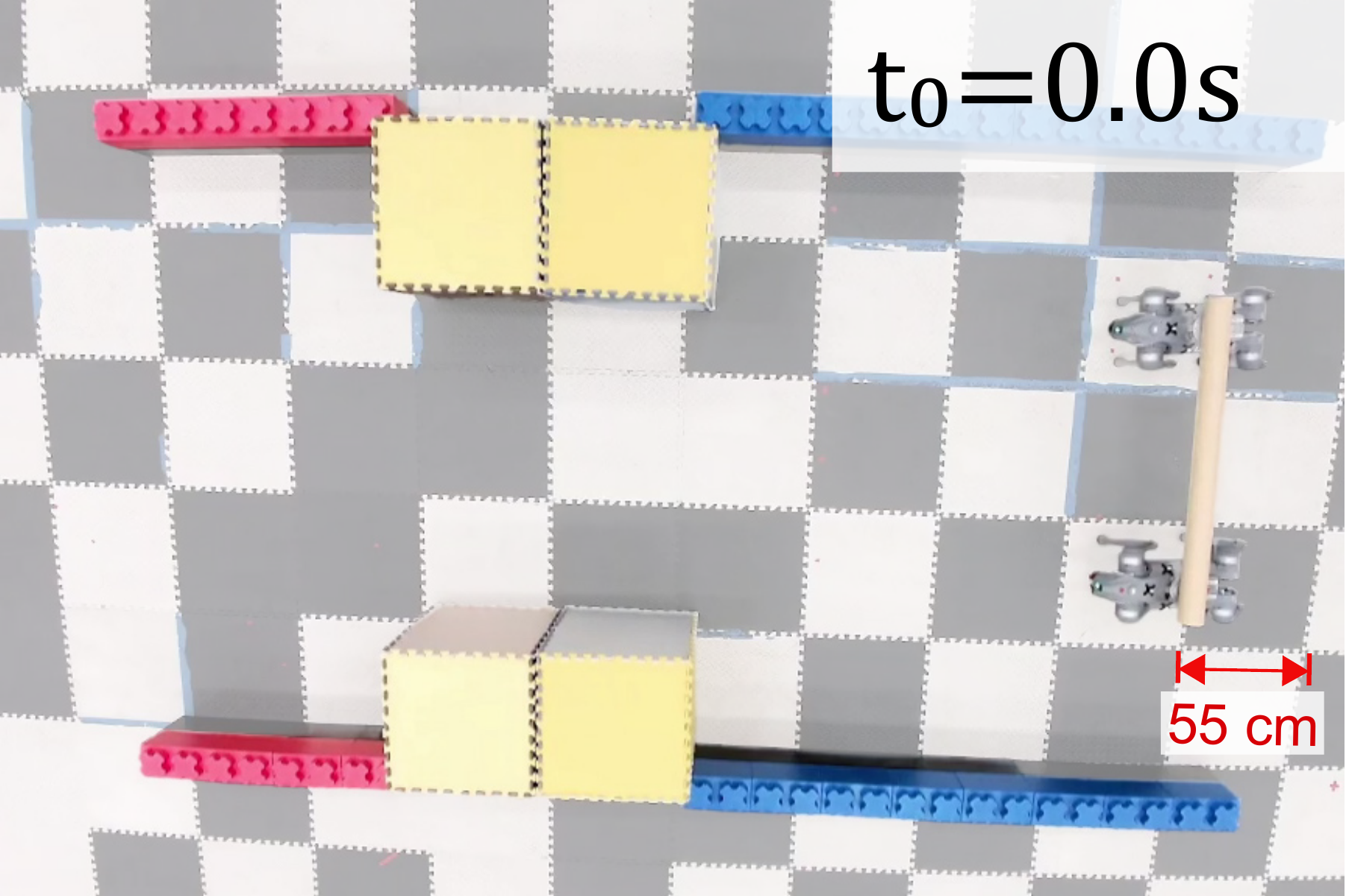}

    \end{minipage}
    \begin{minipage}[b]{0.48\columnwidth}
        \centering
        \includegraphics[width=\linewidth]{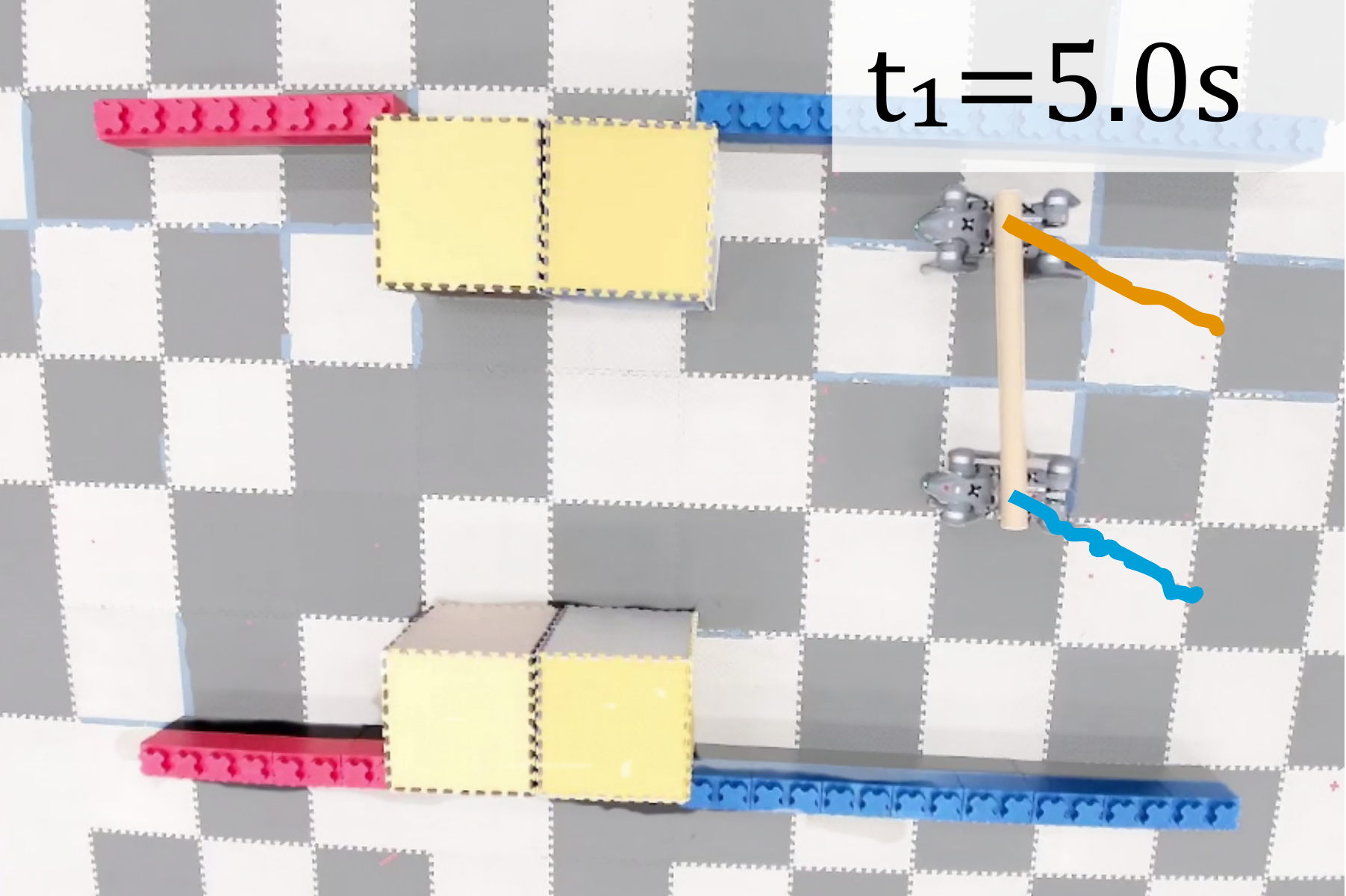}

    \end{minipage}

    \vspace{0.3em} 

    \centering
    \begin{minipage}[b]{0.48\columnwidth}
        \centering
        \includegraphics[width=\linewidth]{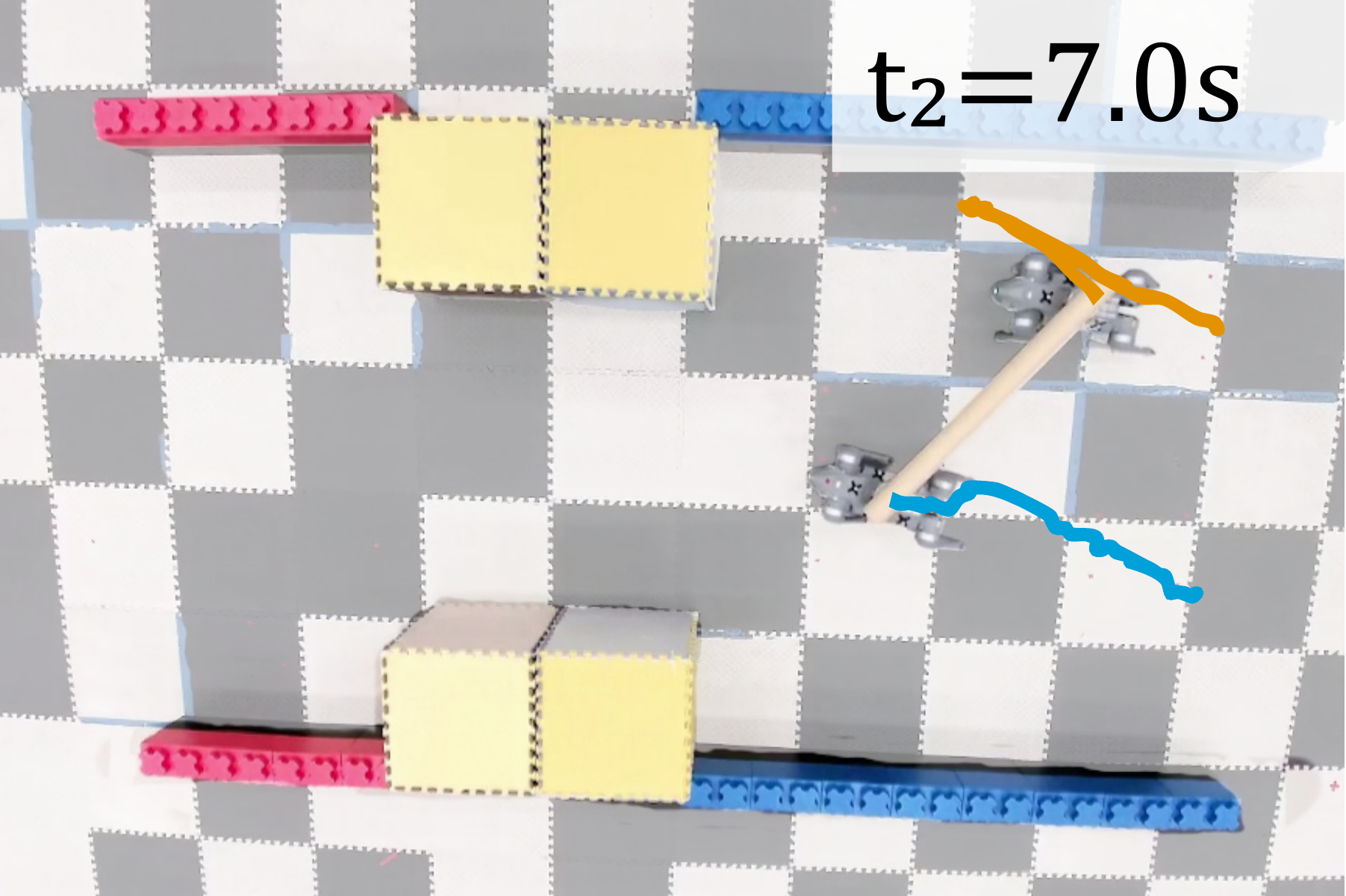}

    \end{minipage}
    \begin{minipage}[b]{0.48\columnwidth}
        \centering
        \includegraphics[width=\linewidth]{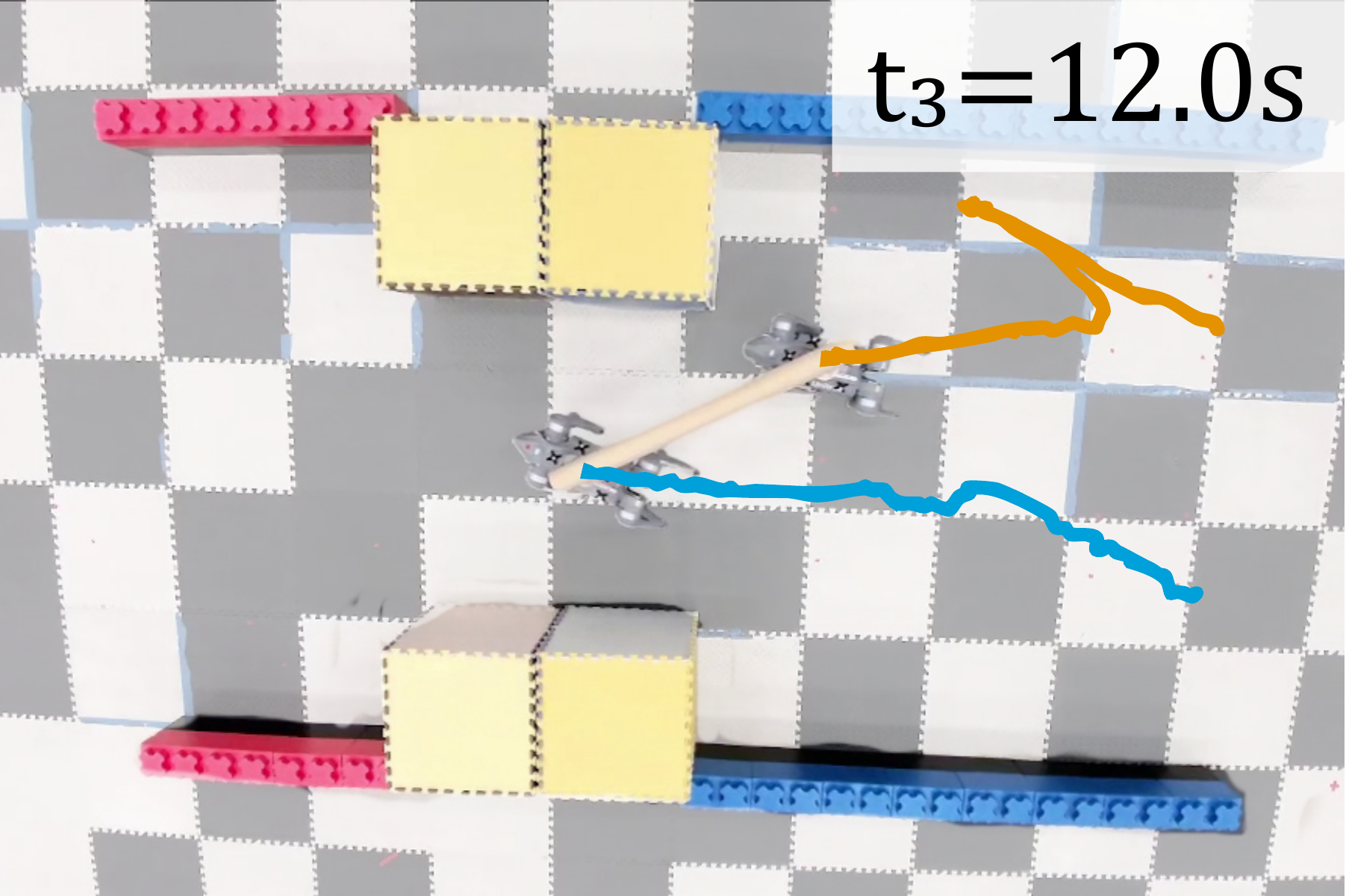}

    \end{minipage}

    \vspace{0.3em} 

    \centering
    \begin{minipage}[b]{0.48\columnwidth}
        \centering
        \includegraphics[width=\linewidth]{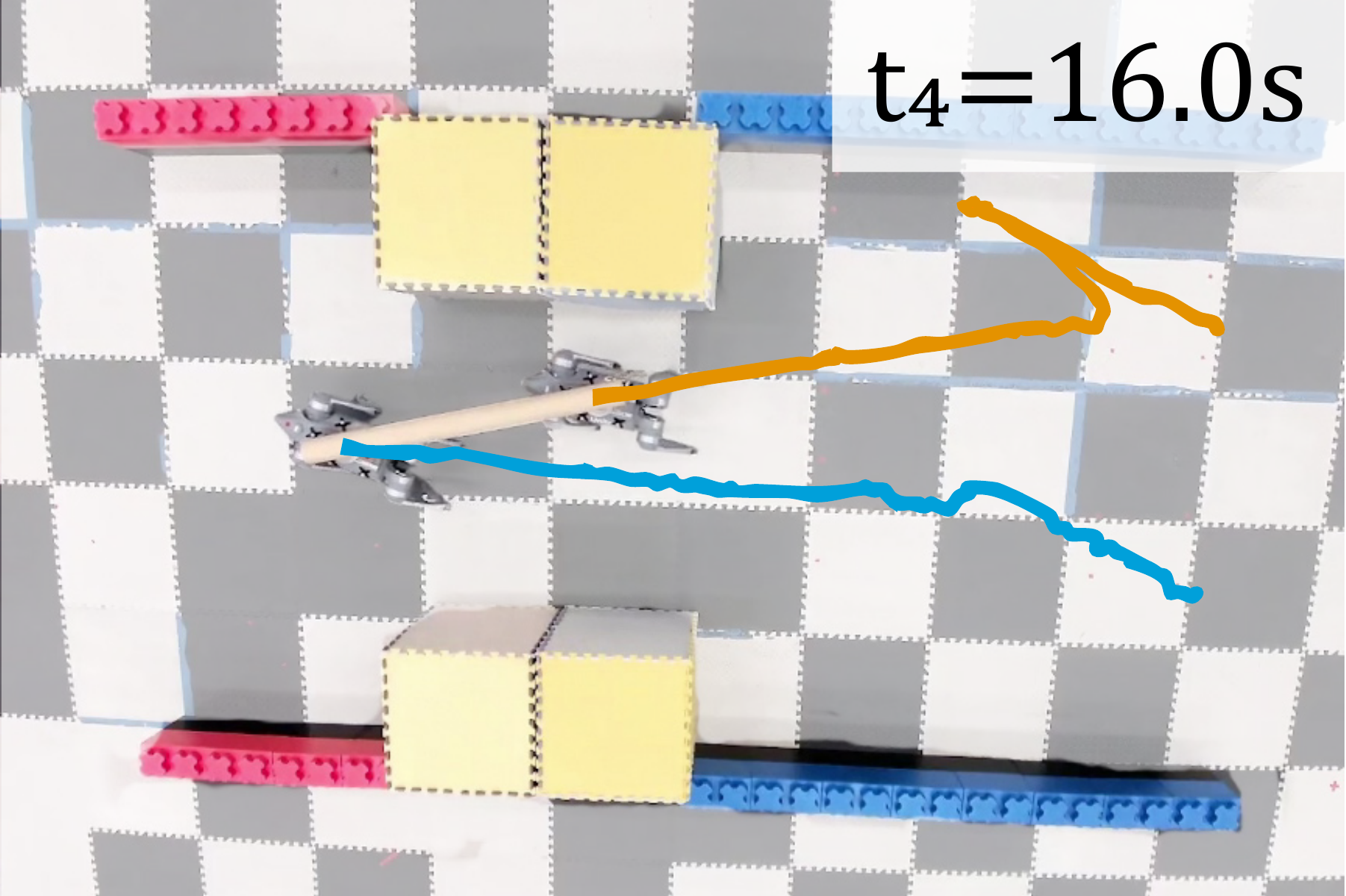}

    \end{minipage}
    \begin{minipage}[b]{0.48\columnwidth}
        \centering
        \includegraphics[width=\linewidth]{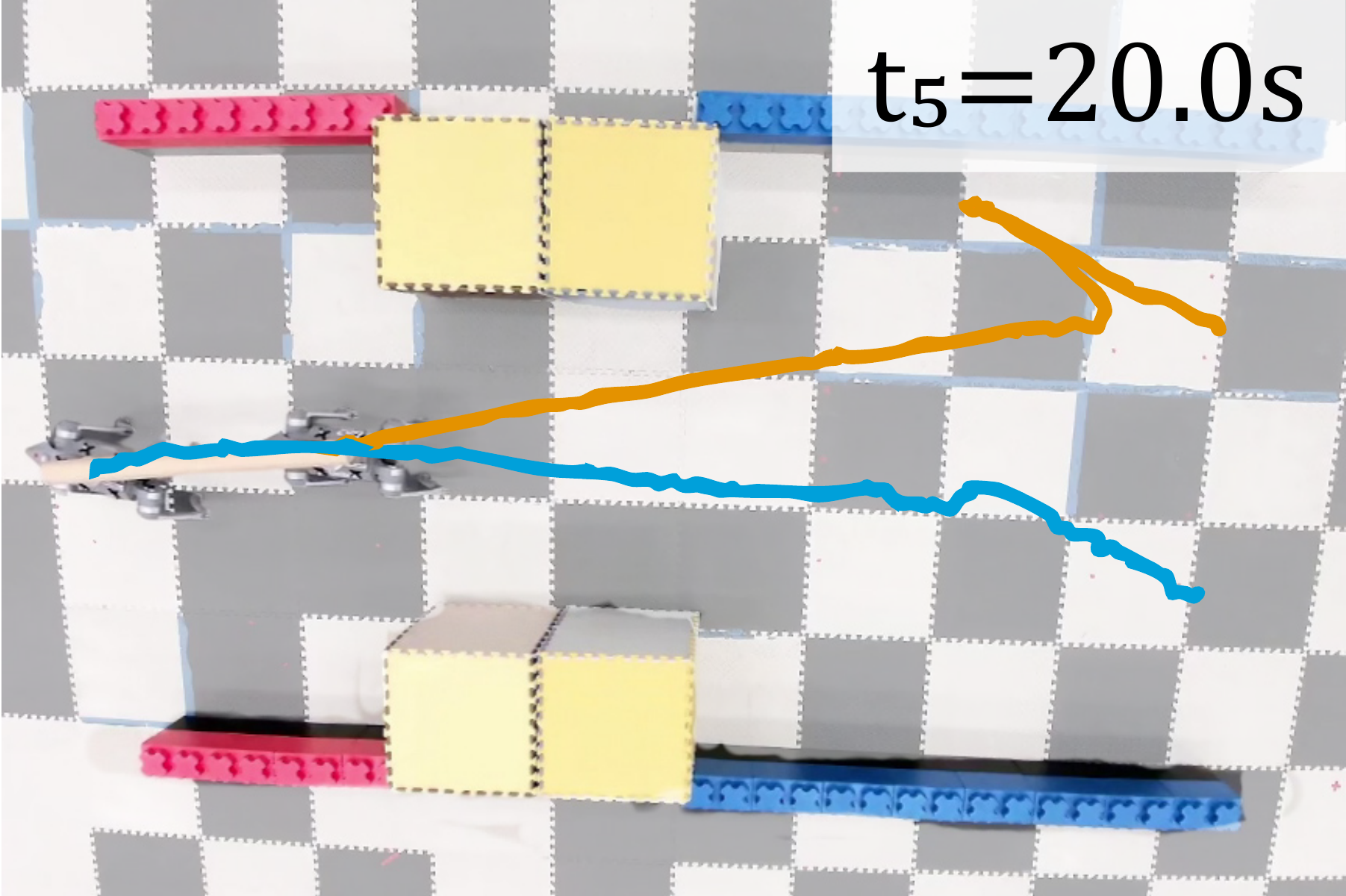}

    \end{minipage}

    \caption{Robot team collaboration for navigating the gate scenario. From $t_1$ to $t_3$, the team collaboratively adjusts its formation, and then passes through the narrow region in a front–rear formation.}
    \label{fig:gate}
\end{figure}

\subsection{Baseline Methods}
To evaluate the performance of our method, we compare it with the following baselines (Table~\ref{tab:baseline}):  

\textbf{HAPPO \cite{kuba2022trust}}: 
In this method, we encode the constraints in~\eqref{original_problem} as penalty terms and integrate them into the reward function as negative rewards.

\textbf{MAPPO \cite{yu2022surprising}}: 
This method also adopts a reward-only design, but all robots share the same parameters instead of maintaining independent policies for each agent.

\textbf{Uniform Constraint Allocation (UCA)}: 
We apply a simple uniform allocation strategy for constraints based on~\eqref{problem2} instead of applying constraints allocation. 

\textbf{MACPO \cite{gu2023safe}}: 
We include MACPO as a baseline for comparison and incorporate the constraint allocation mechanism.

\subsection{Training Performance}
In this section, we conduct a comparative analysis of the training performance of the proposed method against the baseline approaches. 
During training, $300$ environments are used for parallel data collection. 
The weights of the reward and cost functions are set to
$w_f = 1,\; w_d = 10,\; w_m = 1,\; w_c = 5$. 

Fig.~\ref{fig:sim_r_c} shows the evolution of the cost (right) and reward (left) values of each method during training. For the cost part, the penalty terms in HAPPO and MAPPO are accounted for in the cost term. 
Lower cost values indicate better performance in collision avoidance and collaborative motion. 
Our method achieves significantly lower cost than reward-only methods. 
This demonstrates that policies incorporating explicit cost constraints (including our method) provide advantages in terms of safety and collaboration.
For the reward part, our algorithm outperforms other cost-based methods and achieves performance comparable to that of the reward-only method. 
These results indicate that the proposed constraint allocation mechanism is able to encourage effective exploration and convergence to superior task policies while still satisfying safety constraints.


\begin{figure}[t]
    \centering
    \begin{minipage}[b]{0.48\columnwidth}
        \centering
        \includegraphics[width=\linewidth]{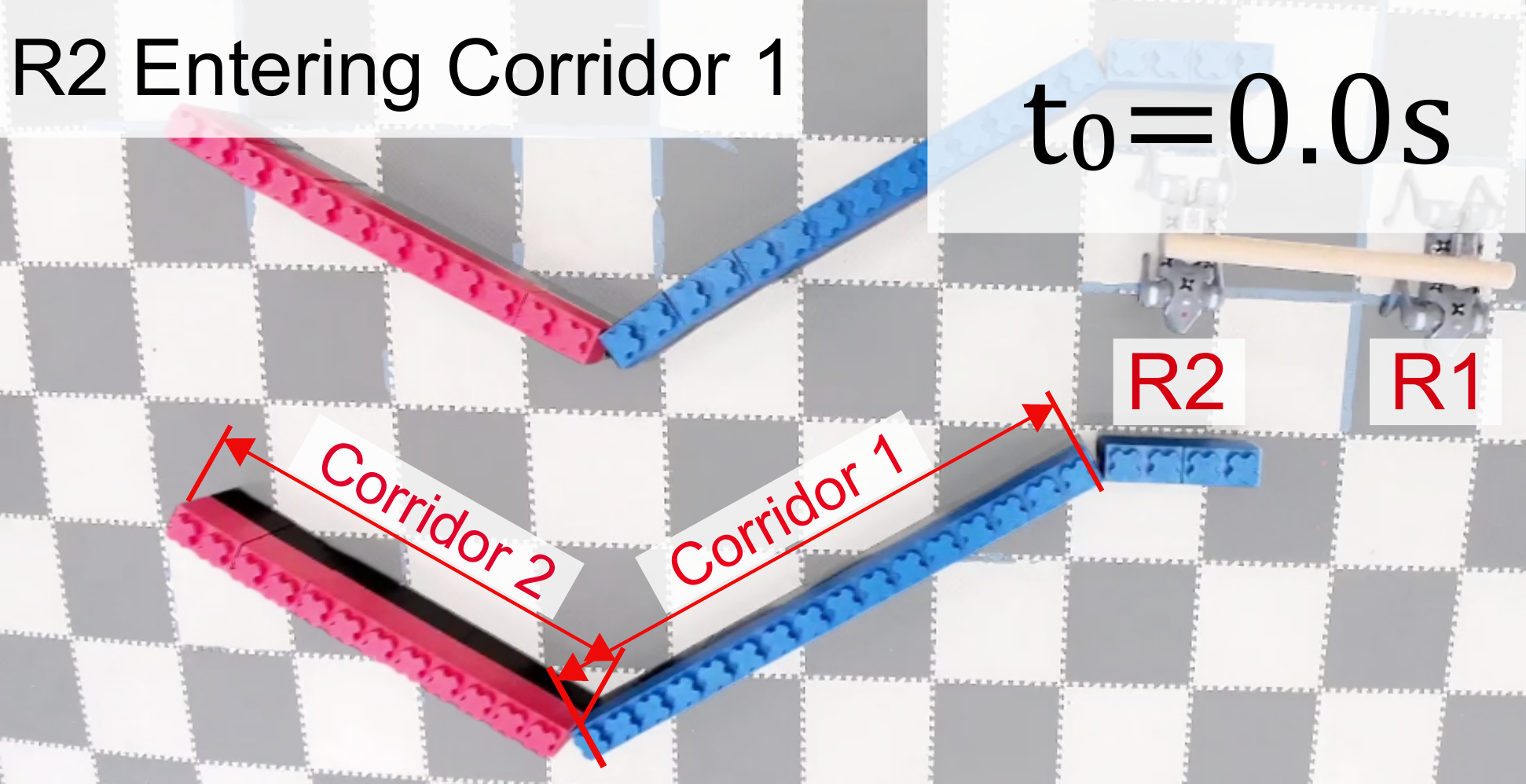}

    \end{minipage}
    \begin{minipage}[b]{0.48\columnwidth}
        \centering
        \includegraphics[width=\linewidth]{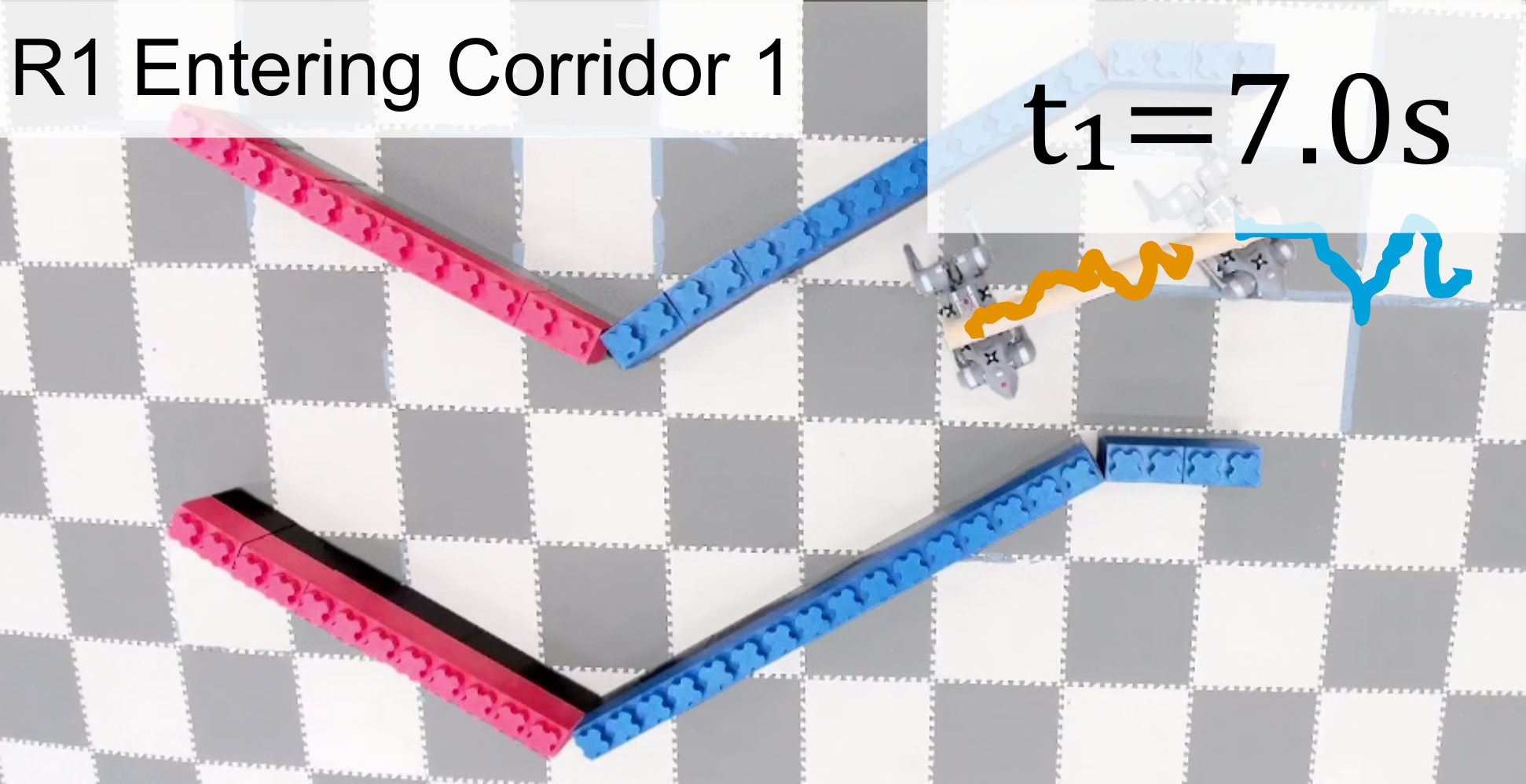}

    \end{minipage}

    \vspace{0.3em} 

    \centering
    \begin{minipage}[b]{0.48\columnwidth}
        \centering
        \includegraphics[width=\linewidth]{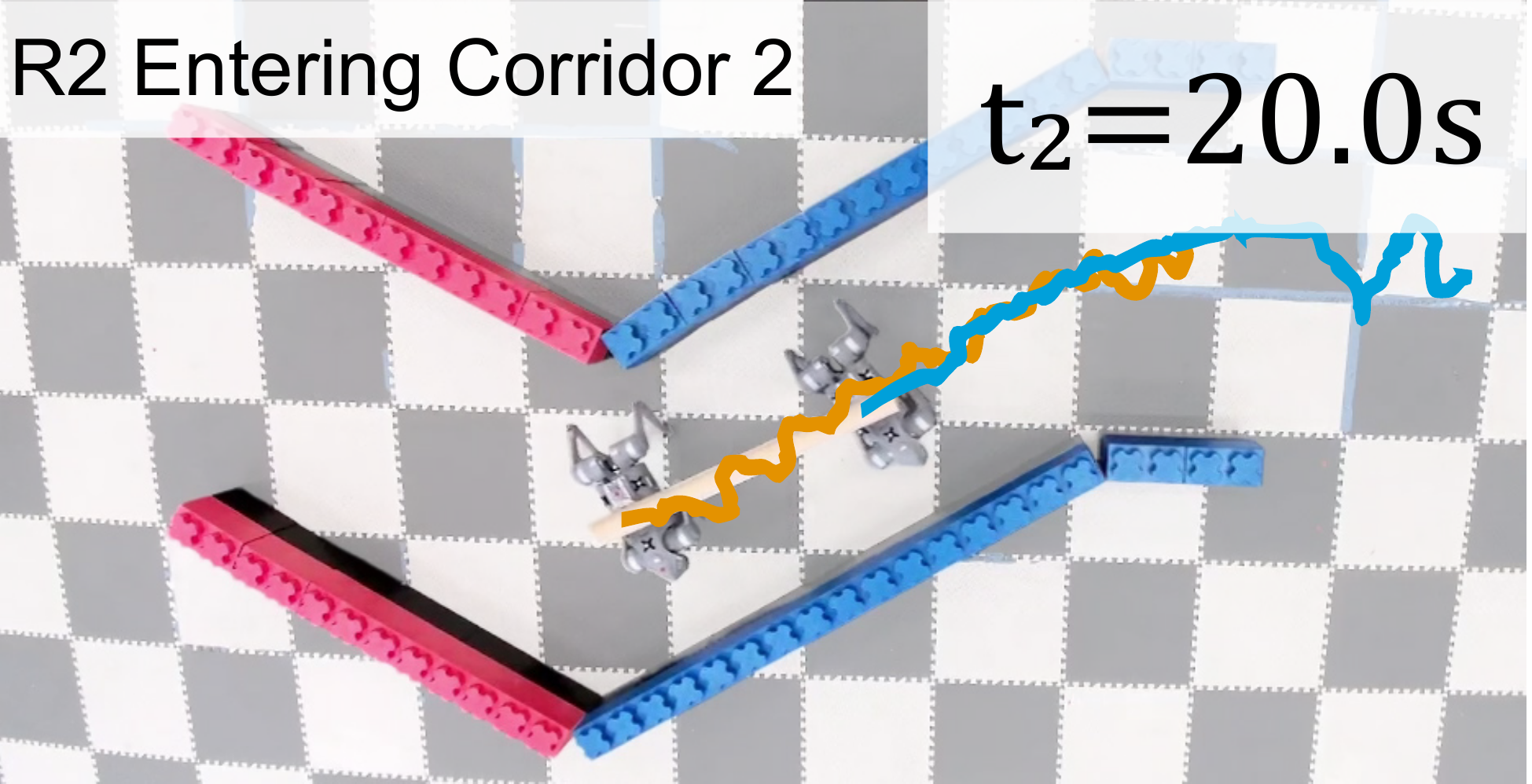}

    \end{minipage}
    \begin{minipage}[b]{0.48\columnwidth}
        \centering
        \includegraphics[width=\linewidth]{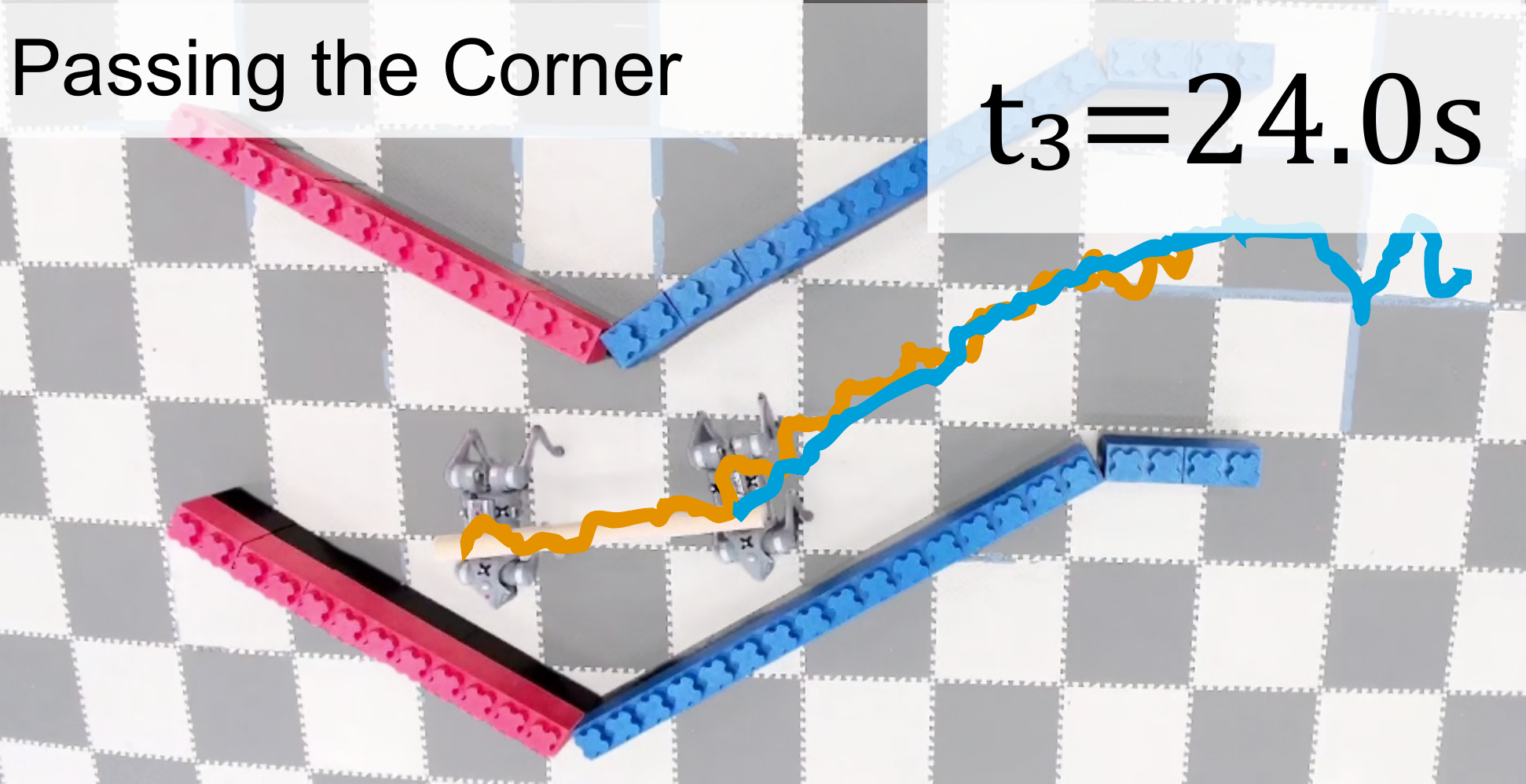}

    \end{minipage}

    \vspace{0.3em} 

    \centering
    \begin{minipage}[b]{0.48\columnwidth}
        \centering
        \includegraphics[width=\linewidth]{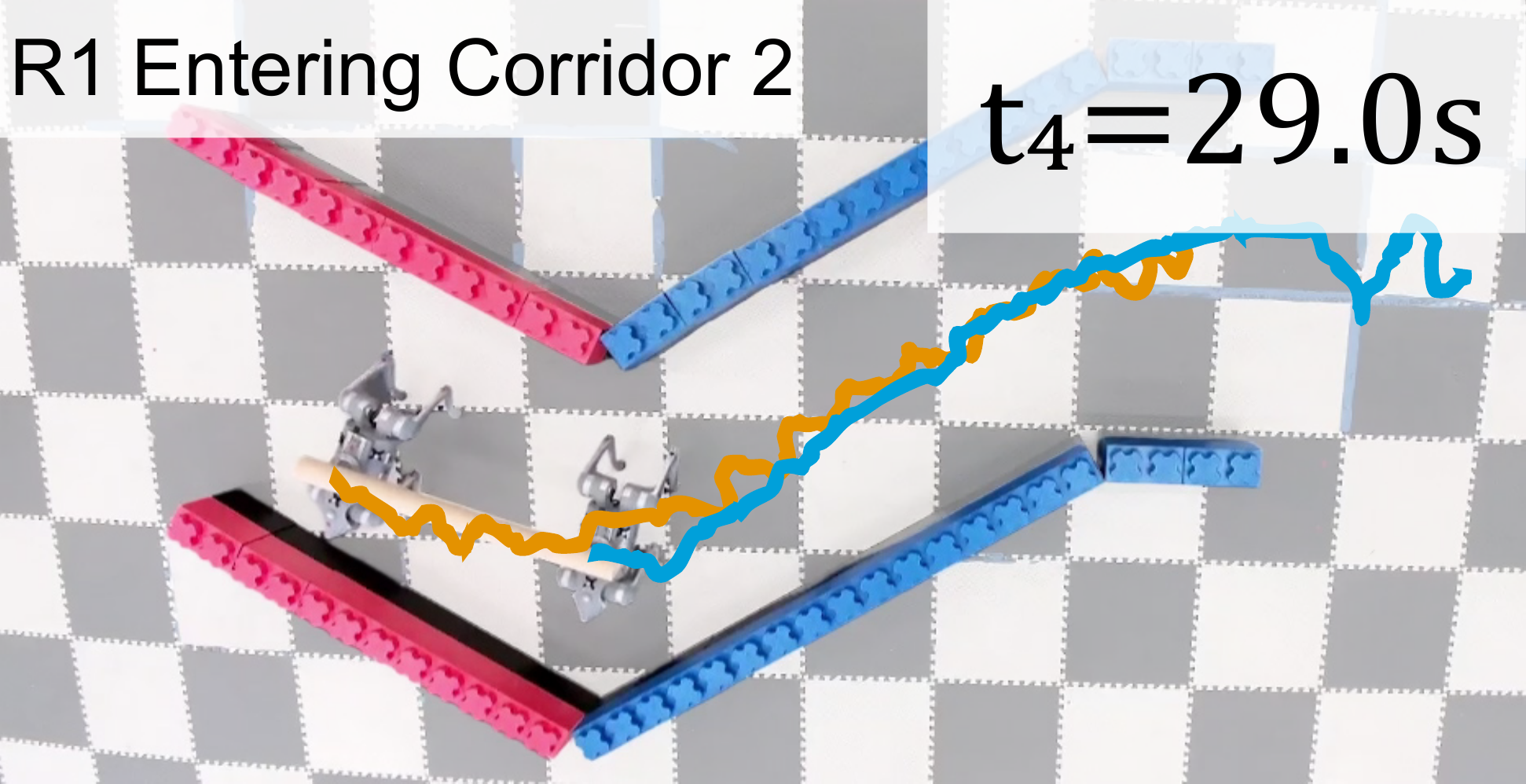}

    \end{minipage}
    \begin{minipage}[b]{0.48\columnwidth}
        \centering
        \includegraphics[width=\linewidth]{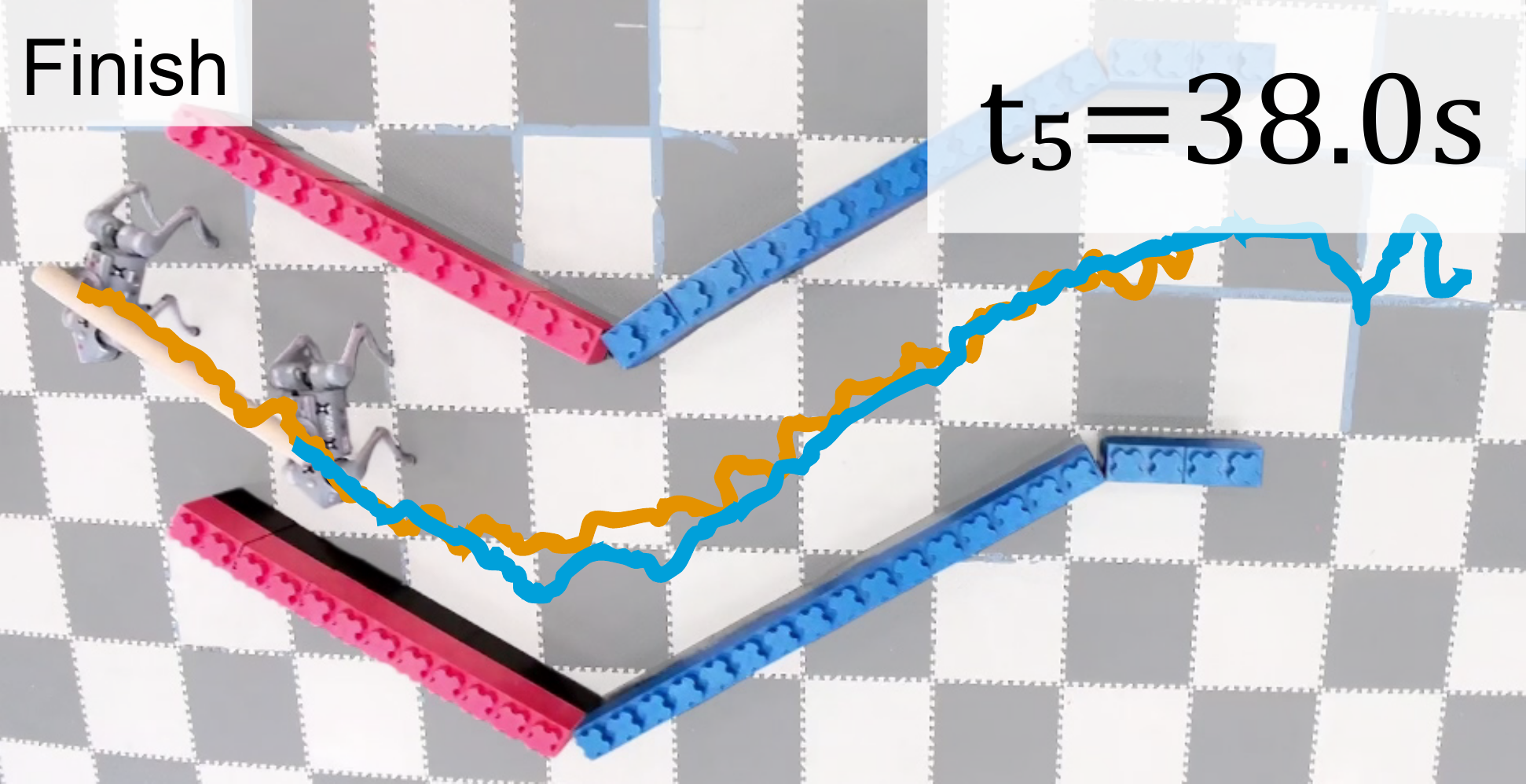}

    \end{minipage}

    \caption{Robot team collaboration in a long and narrow corridor scenario. Robot R1 continuously follows Robot R2 to maintain the formation. From $t_2$ to $t_4$, the team changes its formation to pass through the corner.}
    \label{fig:corridor}
\end{figure}

\subsection{Safety Performance}
In this experiment, we validate the safety performance of the proposed method. 
A total of $30$ independent runs are conducted in the gate scenario, where the initial positions of robots are randomly initialized in each run, 
and the average collision probability per experiment is recorded for each method. 
The results are reported in Table~\ref{tab:safety_gate}. 
Compared with the reward-only method, approaches based on the constrained Markov framework exhibit significantly lower collision probabilities, 
demonstrating superior safety performance. 
Besides, among all constrained Markov methods, our approach achieves the lowest collision probability, indicating the best overall safety performance.

However, an exclusive emphasis on safety may lead to overly conservative policies, resulting in degraded task performance. 
For instance, when the safety penalty weight is large or the environment is highly complex, 
robots may refuse to advance toward the goal and wander locally to avoid collisions. 
Thus, we record the number of trials in which robots both avoid collisions and successfully reach the target to complete the task within a predefined time horizon. 
This metric is used to quantify the task success rate of each method. 
As shown in Table~\ref{tab:safety_gate}, our method achieves a significantly higher success rate than the other approaches, indicating that it maintains safety while effectively mitigating overly conservative behavior.

\subsection{Collaboration Performance}
In this section, we validate the collaboration performance of the proposed method. First, we compare the motion trajectories of the team under different methods. 
Smoother trajectories with shorter path lengths indicate a higher degree of collaboration among robots. 
The results are shown in Fig.~\ref{fig:sim_gate}.
Compared with the results obtained by the UCA method, our approach produces trajectories that are smoother and overall shorter. 
Furthermore, based on the observation that robots with poor collaboration are unable to reach the goal and tend to linger in front of obstacles, resulting in highly curved trajectories, we adopt trajectory straightness as a metric to evaluate collaboration performance (Table~\ref{tab:safety_gate}). Trajectory straightness is defined as the ratio between the distance traveled by the robots along the direction of the goal and the trajectory length. A higher ratio indicates more efficient collaboration behavior.
In addition, we record the consumption time for the robot team to reach the target region. If the robots fail to reach the target within the predefined time horizon, the time is set to the maximum duration of 35~s. As summarized in Table~\ref{tab:safety_gate}, our method achieves the best performance on both trajectory straightness and task completion time among all compared methods.
This result illustrates that the proposed constraint allocation mechanism implicitly captures team-level collaborative behaviors. 
For example, robots with relatively looser constraints possess stronger exploration capabilities during training, 
which facilitates the discovery of more efficient collaborative strategies and ultimately improves overall team performance.

\begin{figure}[t]
    \centering
    \begin{minipage}[b]{0.48\columnwidth}
        \centering
        \includegraphics[width=\linewidth]{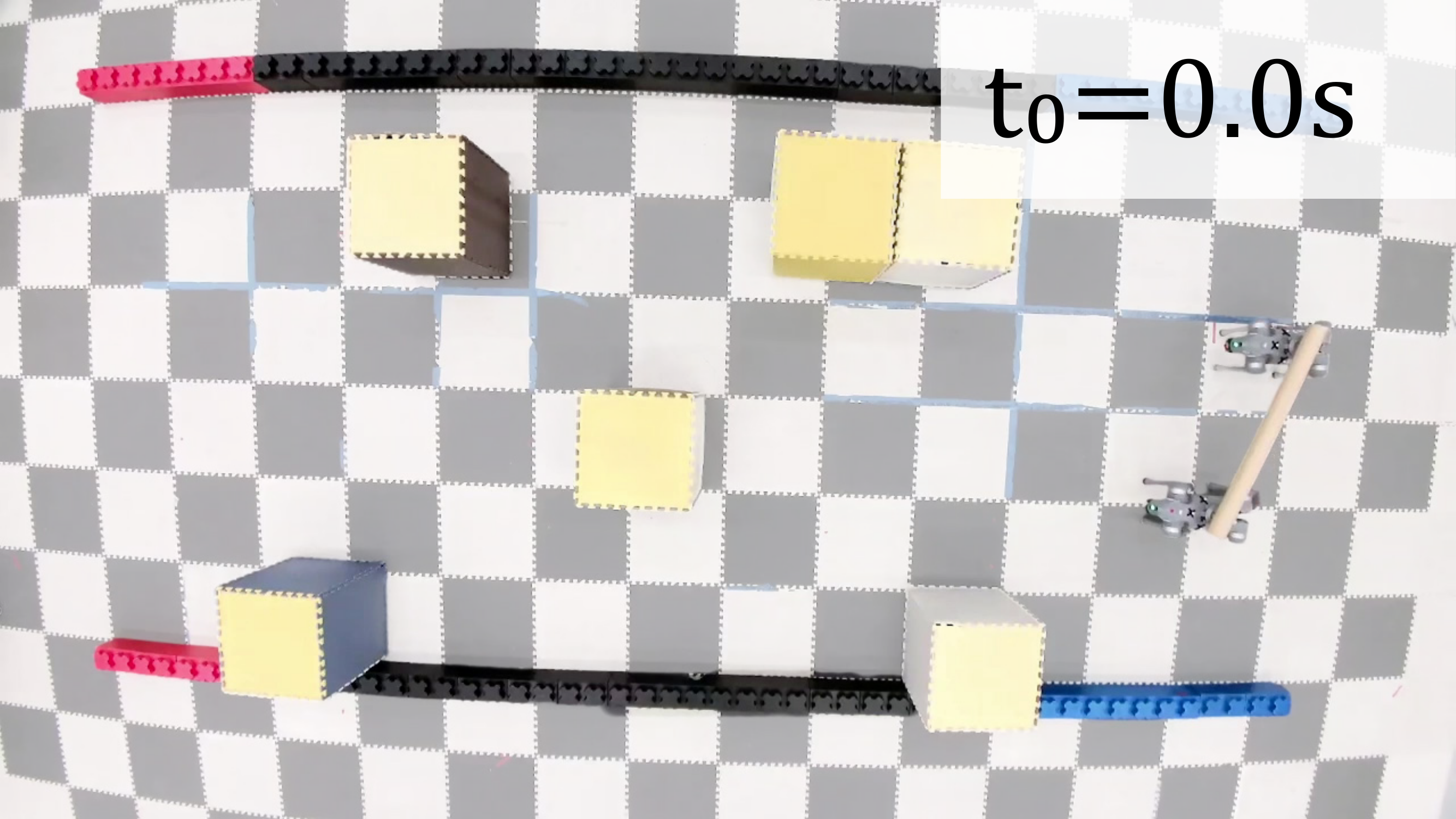}

    \end{minipage}
    \begin{minipage}[b]{0.48\columnwidth}
        \centering
        \includegraphics[width=\linewidth]{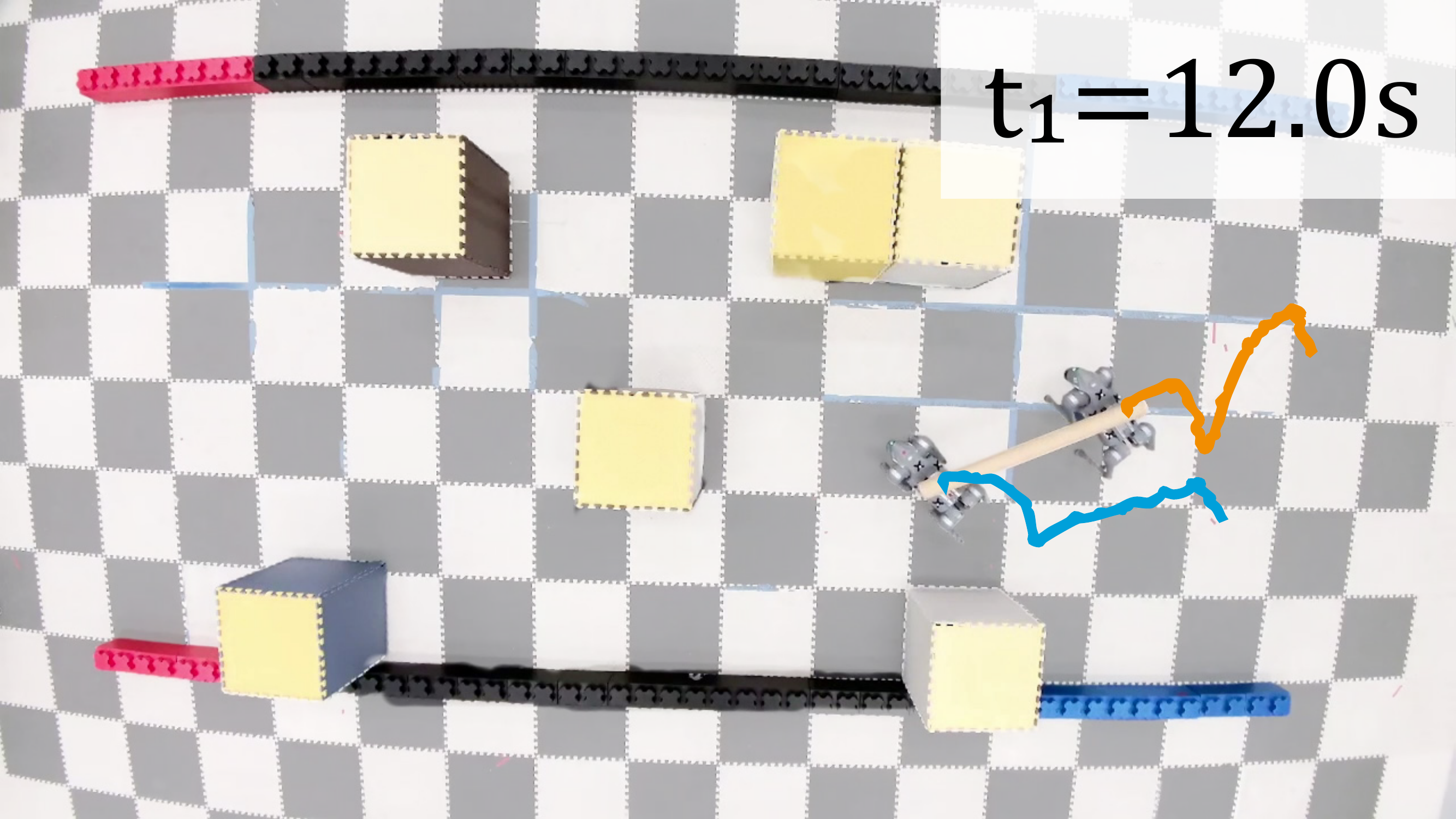}

    \end{minipage}

    \vspace{0.3em} 

    \centering
    \begin{minipage}[b]{0.48\columnwidth}
        \centering
        \includegraphics[width=\linewidth]{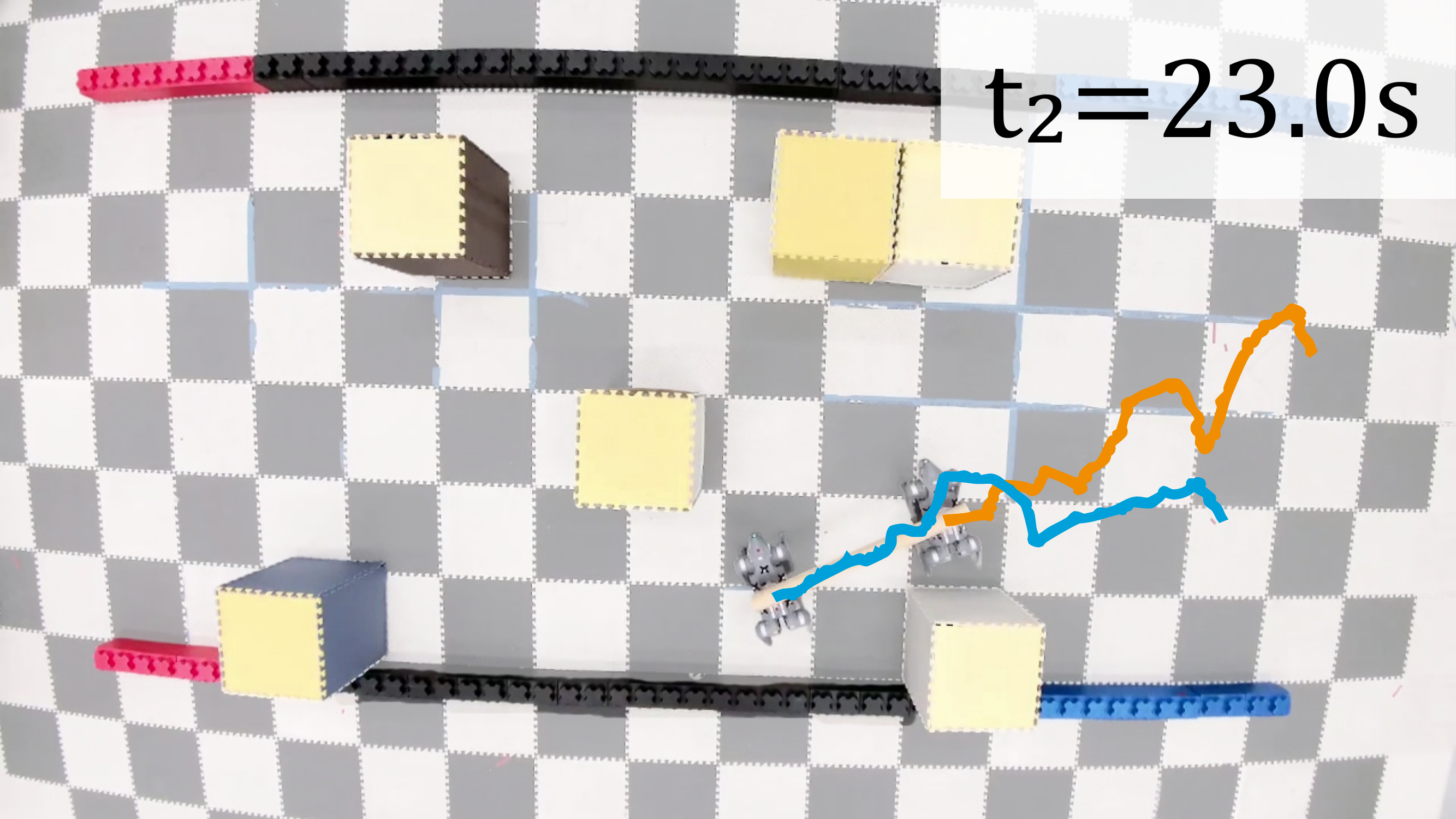}

    \end{minipage}
    \begin{minipage}[b]{0.48\columnwidth}
        \centering
        \includegraphics[width=\linewidth]{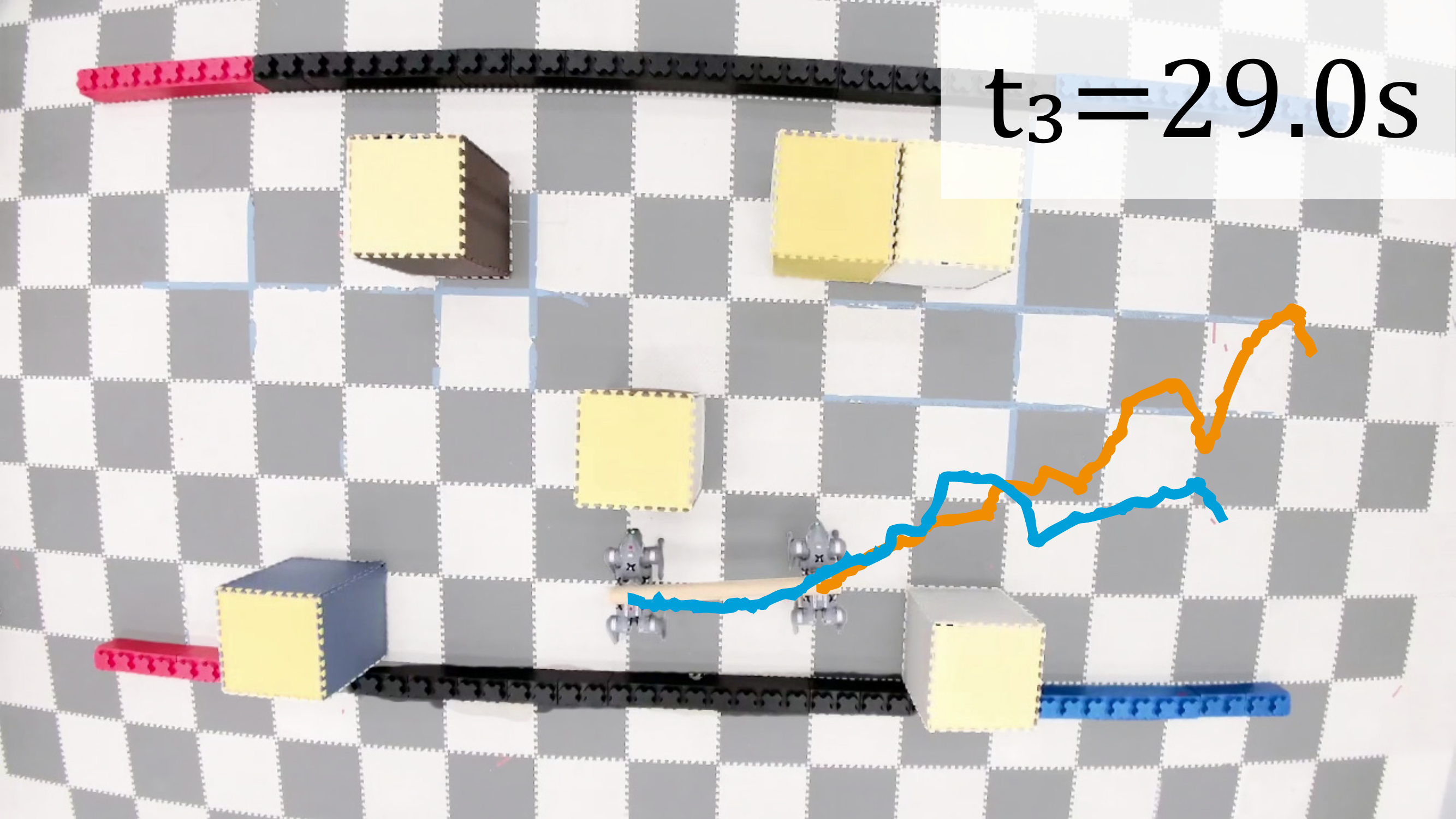}

    \end{minipage}

    \vspace{0.3em} 

    \centering
    \begin{minipage}[b]{0.48\columnwidth}
        \centering
        \includegraphics[width=\linewidth]{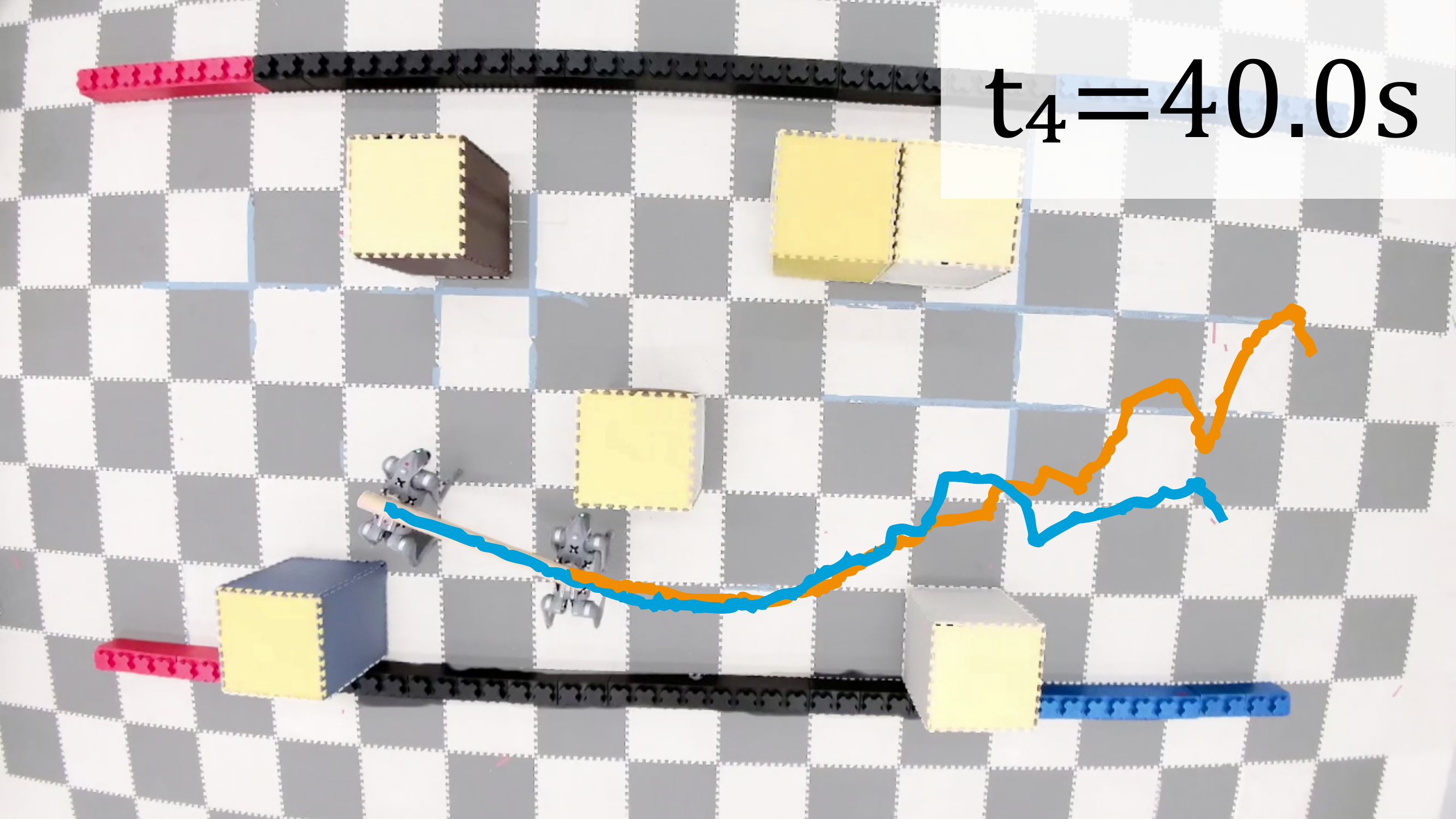}

    \end{minipage}
    \begin{minipage}[b]{0.48\columnwidth}
        \centering
        \includegraphics[width=\linewidth]{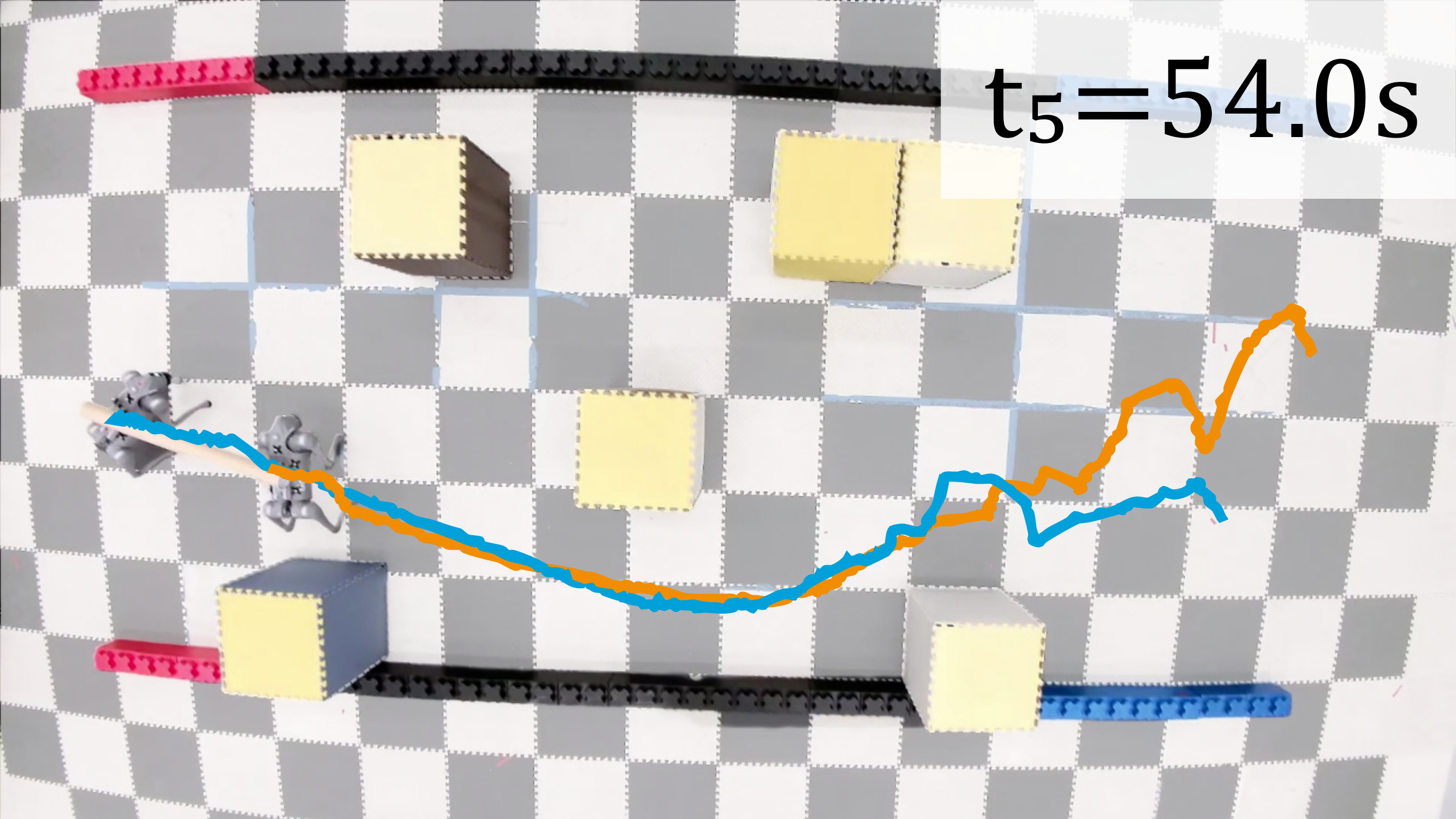}

    \end{minipage}

    \caption{Robot team collaboration in a forest scenario. From $t_1$ to $t_2$, when encountering a junction, the two robots make identical action decisions to avoid conflicts.}
    \label{fig:forest}
\end{figure}

\section{REAL-WORLD EXPERIMENTS}

We design three different experimental scenarios to evaluate the safe and high-performance inter-robot collaboration performance of the proposed method in real-world environments. 
In the experiments, the carried object has a length of $1.2$~m and a diameter of $0.1$~m, 
and it is rigidly fixed to the rotating platforms mounted on two Go2 robots. 
A motion capture team is employed to obtain the pose information of robots in the world coordinate frame.

\subsection{Gate Scenario}
This experiment aims to evaluate the formation transformation capability of the team. 
To increase the task difficulty, compared with the simulation setup, we reduce the gate width to $1.2$~m and increase its depth to $1.2$~m.
The result is illustrated in Fig.~\ref{fig:gate}. 
Our method is able to adaptively transform the initially side-by-side formation into a front--rear formation before entering the gate, 
thereby enabling stable and smooth passage. 
The experimental results demonstrate that, under the proposed method, the robot team is able to adaptively change its formation according to the environmental context and collaboratively navigate through complex scenarios.

\subsection{Corridor Scenario}
This experiment is designed to evaluate long-term safety guarantees of the team in narrow environments. 
The experimental scenario consists of two straight corridors, each with a length of $2.7$~m and a width of $1.3$~m. 
Moreover, to evaluate collaborative capability of the team in narrow environments, we design a connection between two corridors by a corner with an angle of $117$ degrees. 
The result is illustrated in Fig.~\ref{fig:corridor}. 
Our method is able to maintain a stable formation while navigating through the long, narrow corridors and effectively avoid collisions. 
The experimental results demonstrate that, under the proposed method, the robot team is capable of maintaining its formation over extended periods, while also performing formation reconfiguration in confined spaces.

\subsection{Forest Scenario}
This experiment is designed to evaluate the collaborative decision-making capability of a robot team when encountering multiple branching intersections. 
Obstacles are sparsely distributed in an environment with a length of $7.2$~m and a width of $3.3$~m, forming several branching junctions. 
The trajectory results of the robot team navigating the environment are shown in Fig.~\ref{fig:forest}. 
When approaching branching intersections, the robots in the team are able to simultaneously make consistent forward movement decisions, thereby avoiding conflicts.

%% file: 5Conclusion/conclusion.tex
\section{CONCLUSION}

In this work, we propose a novel RL framework to address safe and high-performance inter-robot collaboration challenges in decentralized transportation systems. 
The framework introduces a constraint allocation mechanism that distributes the constraint terms in a constrained Markov decision game among team members, implicitly encouraging agents to assume different roles and collaboratively accomplish the task. 
Through extensive simulations and real-world experiments, we demonstrate that the proposed framework significantly improves both the safety and performance of collaborative behavior. 
The robot team is able to flexibly reconfigure its formation in narrow environments and successfully transport a payload to the target location.

%% file: ref.bib
@article{kim2023layered,
  title={Layered control for cooperative locomotion of two quadrupedal robots: Centralized and distributed approaches},
  author={Kim, Jeeseop and Fawcett, Randall T and Kamidi, Vinay R and Ames, Aaron D and Hamed, Kaveh Akbari},
  journal={IEEE Transactions on Robotics},
  volume={39},
  number={6},
  pages={4728--4748},
  year={2023},
  publisher={IEEE}
}

@inproceedings{shibata2021deep,
  title={Deep reinforcement learning of event-triggered communication and control for multi-agent cooperative transport},
  author={Shibata, Kazuki and Jimbo, Tomohiko and Matsubara, Takamitsu},
  booktitle={2021 IEEE International Conference on Robotics and Automation},
  pages={8671--8677},
  year={2021}
}

@inproceedings{ji2021reinforcement,
  title={Reinforcement learning for collaborative quadrupedal manipulation of a payload over challenging terrain},
  author={Ji, Yandong and Zhang, Bike and Sreenath, Koushil},
  booktitle={2021 IEEE 17th International Conference on Automation Science and Engineering},
  pages={899--904},
  year={2021}
}

@article{lei2025safe,
  title={Safe Motion Planning for Multi-Vehicle Autonomous Driving in Uncertain Environment},
  author={Lei, Zhezhi and Wang, Wenxin and Zhu, Zicheng and Ma, Jun and Ge, Shuzhi Sam},
  journal={IEEE Robotics and Automation Letters},
  year={2025},
  publisher={IEEE},
  volume={10},
  pages={2199-2206},
}

@inproceedings{margolis2023walk,
  title={Walk these ways: Tuning robot control for generalization with multiplicity of behavior},
  author={Margolis, Gabriel B and Agrawal, Pulkit},
  booktitle={Conference on Robot Learning},
  pages={22--31},
  year={2023}
}

@article{krizmancic2020cooperative,
  title={Cooperative aerial-ground multi-robot system for automated construction tasks},
  author={Krizmancic, Marko and Arbanas, Barbara and Petrovic, Tamara and Petric, Frano and Bogdan, Stjepan},
  journal={IEEE Robotics and Automation Letters},
  volume={5},
  number={2},
  pages={798--805},
  year={2020},
  publisher={IEEE}
}

@article{nie2024social,
  title={Social-Learning Coordination of Collaborative Multi-Robot Systems Achieves Resilient Production in a Smart Factory},
  author={Nie, Zixiang and Chen, Kwang-Cheng and Kim, Kyeong Jin},
  journal={IEEE Transactions on Automation Science and Engineering},
  year={2024},
  publisher={IEEE}
}

@inproceedings{schulman2015trust,
  title={Trust region policy optimization},
  author={Schulman, John and Levine, Sergey and Abbeel, Pieter and Jordan, Michael and Moritz, Philipp},
  booktitle={International conference on machine learning},
  pages={1889--1897},
  year={2015}
}

@article{yu2022surprising,
  title={The surprising effectiveness of {PPO} in cooperative multi-agent games},
  author={Yu, Chao and Velu, Akash and Vinitsky, Eugene and Gao, Jiaxuan and Wang, Yu and Bayen, Alexandre and Wu, Yi},
  journal={Advances in neural information processing systems},
  volume={35},
  pages={24611--24624},
  year={2022}
}

@inproceedings{kuba2022trust,
  title={Trust Region Policy Optimisation in Multi-Agent Reinforcement Learning},
  author={Kuba, JG and Chen, R and Wen, M and Wen, Y and Sun, F and Wang, J and Yang, Y},
  booktitle={ICLR 2022-10th International Conference on Learning Representations},
  note = {pp. 1046},
  year={2022}
}

@article{cao2024hma,
  title={Hma-sar: Multi-agent search and rescue for unknown located dynamic targets in completely unknown environments},
  author={Cao, Xiao and Li, Mingyang and Tao, Yuting and Lu, Peng},
  journal={IEEE Robotics and Automation Letters},
  volume={9},
  number={6},
  pages={5567--5574},
  year={2024},
  publisher={IEEE}
}

@article{gu2023safe,
  title={Safe multi-agent reinforcement learning for multi-robot control},
  author={Gu, Shangding and Kuba, Jakub Grudzien and Chen, Yuanpei and Du, Yali and Yang, Long and Knoll, Alois and Yang, Yaodong},
  journal={Artificial Intelligence},
  volume={319},
  note = {pp. 103905},
  year={2023},
  publisher={Elsevier}
}

@article{sui2020formation,
  title={Formation control with collision avoidance through deep reinforcement learning using model-guided demonstration},
  author={Sui, Zezhi and Pu, Zhiqiang and Yi, Jianqiang and Wu, Shiguang},
  journal={IEEE Transactions on Neural Networks and Learning Systems},
  volume={32},
  number={6},
  pages={2358--2372},
  year={2020},
  publisher={IEEE}
}

@article{hou2023multiagent,
  title={A multiagent cooperative learning system with evolution of social roles},
  author={Hou, Yaqing and Sun, Mingyang and Zeng, Yifeng and Ong, Yew-Soon and Jin, Yaochu and Ge, Hongwei and Zhang, Qiang},
  journal={IEEE Transactions on Evolutionary Computation},
  volume={28},
  number={2},
  pages={531--543},
  year={2023},
  publisher={IEEE}
}

@article{han2020actor,
  title={Actor-critic reinforcement learning for control with stability guarantee},
  author={Han, Minghao and Zhang, Lixian and Wang, Jun and Pan, Wei},
  journal={IEEE Robotics and Automation Letters},
  volume={5},
  number={4},
  pages={6217--6224},
  year={2020},
  publisher={IEEE}
}

@article{jose2024bilevel,
  title={Bilevel Learning for Dual-Quadruped Collaborative Transportation under Kinematic and Anisotropic Velocity Constraints},
  author={Jose, Williard Joshua and Zhang, Hao},
  journal={arXiv preprint arXiv:2412.08644},
  year={2024}
}

@inproceedings{feng2025learning,
  title={Learning multi-agent loco-manipulation for long-horizon quadrupedal pushing},
  author={Feng, Yuming and Hong, Chuye and Niu, Yaru and Liu, Shiqi and Yang, Yuxiang and Zhao, Ding},
  booktitle={2025 IEEE International Conference on Robotics and Automation},
  pages={14441--14448},
  year={2025}
}

@article{tian2025multi,
  title={Multi-Robot Cooperative Transportation of Irregular Objects by Multi-Objective Optimization With Distributed Control},
  author={Tian, Dujie and Zhou, Lelai and Zhang, Chen and Li, Yibin},
  journal={IEEE Robotics and Automation Letters},
  year={2025},
  publisher={IEEE}
}

@article{nie2024predictive,
  title={Predictive path coordination of collaborative transportation multirobot system in a smart factory},
  author={Nie, Zixiang and Chen, Kwang-Cheng},
  journal={IEEE Transactions on Systems, Man, and Cybernetics: Systems},
  year={2024},
  publisher={IEEE}
}

@inproceedings{sundin2022decentralized,
  title={Decentralized model predictive control for equilibrium-based collaborative uav bar transportation},
  author={Sundin, Roberto C and Roque, Pedro and Dimarogonas, Dimos V},
  booktitle={2022 International Conference on Robotics and Automation},
  pages={4915--4921},
  year={2022}
}

@inproceedings{grannen2023stabilize,
  title={Stabilize to act: Learning to coordinate for bimanual manipulation},
  author={Grannen, Jennifer and Wu, Yilin and Vu, Brandon and Sadigh, Dorsa},
  booktitle={Conference on Robot Learning},
  pages={563--576},
  year={2023}
}

@inproceedings{lin2024projection,
  title={Projection-Based Fast and Safe Policy Optimization for Reinforcement Learning},
  author={Lin, Shijun and Wang, Hao and Chen, Ziyang and Kan, Zhen},
  booktitle={2024 IEEE International Conference on Robotics and Automation},
  pages={7426--7432},
  year={2024},
  organization={IEEE}
}

@article{razmjoo2024sampling,
  title={Sampling-based constrained motion planning with products of experts},
  author={Razmjoo, Amirreza and Xue, Teng and Shetty, Suhan and Calinon, Sylvain},
  journal={The International Journal of Robotics Research},
  note = {pp. 02783649251404955},
  year={2024},
  publisher={SAGE Publications Sage UK: London, England}
}

@article{imran2025safety,
  title={Safety-critical and distributed nonlinear predictive controllers for teams of quadrupedal robots},
  author={Imran, Basit Muhammad and Kim, Jeeseop and Chunawala, Taizoon and Leonessa, Alexander and Hamed, Kaveh Akbari},
  journal={IEEE Robotics and Automation Letters},
  year={2025},
  publisher={IEEE}
}
